\documentclass[10pt]{article} 

\usepackage{amsmath,amsfonts,bm}

\def\eqref#1{equation~\ref{#1}}

\def\1{\bm{1}}

\DeclareMathAlphabet{\mathsfit}{\encodingdefault}{\sfdefault}{m}{sl}
\SetMathAlphabet{\mathsfit}{bold}{\encodingdefault}{\sfdefault}{bx}{n}

\usepackage{amsmath}
\usepackage{amssymb}
\usepackage{mathtools}
\usepackage{amsthm}
\usepackage{stackengine}
\usepackage{hyperref}
\usepackage{url}
\usepackage{microtype}
\usepackage{graphicx}
\usepackage{algorithm}
\usepackage{algorithmic}
\usepackage{subfigure}
\usepackage{subcaption}

\usepackage[accepted,preprint]{tmlr}

\newtheorem{theorem}{Theorem}[section]

\newtheorem{lemma}[theorem]{Lemma}

\theoremstyle{definition}
\newtheorem{definition}[theorem]{Definition}

\theoremstyle{remark}

\def\delequal{\mathrel{\ensurestackMath{\stackon[1pt]{=}{\scriptstyle\Delta}}}}

\title{Extended Deep Submodular Functions}

\author{\name Seyed Mohammad Hosseini* \email semo.hosseini@sharif.edu \\
      \addr Department of Computer Engineering\\
      Sharif University of Technology
      \ANDD
      \name Arash Jamshidi* \email arashjamshidi@sharif.edu \\
      \addr Department of Computer Engineering\\
      Sharif University of Technology
      \ANDD
      \name Seyed Mahdi Noormousavi \email mahdinoormousavi75@sharif.edu\\
      \addr Department of Computer Engineering\\
      Sharif University of Technology
      \ANDD
      \name Mahdi Jafari Siavoshani \email mjafari@sharif.edu \\
      \addr Department of Computer Engineering\\
      Sharif University of Technology
      \ANDD
      \name Naeimeh Omidvar \email omidvar@ipm.ir \\
      \addr Institute for Research in Fundamental Sciences (IPM)
      }

\def\openreview{\url{https://openreview.net/forum?id=XXXX}} 

\begin{document}

\maketitle

\begin{abstract}
We introduce a novel category of set functions called Extended Deep Submodular functions (EDSFs), which are neural network-representable. EDSFs serve as an extension of Deep Submodular Functions (DSFs), inheriting crucial properties from DSFs while addressing innate limitations. It is known that DSFs can represent a limiting subset of submodular functions. In contrast, through an analysis of polymatroid properties, we establish that EDSFs possess the capability to represent all monotone submodular functions, a notable enhancement compared to DSFs. Furthermore, our findings demonstrate that EDSFs can represent any monotone set function, indicating the family of EDSFs is equivalent to the family of all monotone set functions. Additionally, we prove that EDSFs maintain the concavity inherent in DSFs when the components of the input vector are non-negative real numbers—an essential feature in certain combinatorial optimization problems. Through extensive  experiments, we illustrate that EDSFs exhibit significantly lower empirical generalization error than DSFs in the learning of coverage functions. This suggests that EDSFs present a promising advancement in the representation and learning of set functions with improved generalization capabilities.
\end{abstract}

\section{Introduction}

\label{sec:intro}
Submodular functions have found extensive applications in various fields of study, including modeling influence in social networks \cite{kempe2003maximizing}, energy functions in probabilistic models \cite{gillenwater2012near}, and clustering \cite{narasimhan2005q}. In particular in economics, a wide range of scenarios incorporate the concept of diminishing marginal return, where acquiring more goods results in diminishing the overall satisfaction or the so-called ``utility'' \cite{mclaughlin2001health,kimball2024diminishing}.

There are several challenges associated with the widespread use of these functions in recent machine learning applications. To improve the modeling of submodular functions in these applications, efforts have been made to represent submodular functions using differentiable functions, such as neural networks. This representation enables the solution of key submodular optimization problems through the utilization of gradient-based methods and convex optimization. 
To illustrate this, suppose we want to maximize a submodular function $f$ under certain constraints. In the exhaustive-search approach, we would have to check all $2^n$ subsets, which is exponential in the size of the ground set $S$. However, with a differentiable representation of $f$, we can utilize first-order information of the function, such as gradients, and project onto the constraint set to solve the optimization problem more efficiently.

There have been previous efforts to represent submodular functions using neural networks. For instance, in \cite{bilmes_deep_2017}, the authors introduced a neural network architecture called Deep Submodular Functions (DSFs), consisting of feedforward layers with non-negative weights and normalized non-decreasing concave activation functions. Functions in this class exhibit interesting properties, such as the concavity of the function when the components of the input vector are all non-negative real numbers. By increasing the number of layers in the DSFs architecture, the family expands, indicating that there are functions in DSFs with $n+1$ layers that are not present in DSFs with $n$ layers. However, as stated by the authors, DSFs cannot represent all monotone submodular functions, which highly restricts their applicability to many machine learning problems.

In this paper, we introduce a novel neural network architecture, called Extended Deep Submodular Functions (EDSFs), which not only have the capability to represent any monotone submodular functions but can also represent any monotone set functions. Moreover, same as in DSFs, when the components of the input vector are all non-negative real numbers, EDSFs are concave, an important feature applicable in various combinatorial optimization settings. In addition, our experiments demonstrate that EDSFs are able to learn one of the most complicated monotone submodular functions, i.e., coverage functions, with significantly lower empirical generalization error compared to DSFs.

The rest of the paper is organized as follows: In Section 2, we formally define DSFs and state some of their important properties. In Section 3, we introduce our augmented architecture to represent all monotone (submodular) set functions and provide a proof. In Section 4, we demonstrate the superior performance of EDSFs in learning coverage functions through numerical evaluations.

\subsection{Related Works}
The exploration of neural networks for modeling submodular functions is relatively sparse in the existing literature. A notable contribution is the introduction of DSFs \cite{bilmes_deep_2017,dolhansky2016deep}. Building on this work, the authors in  \cite{bai_submodular_2018} address the maximization of DSFs under matroid constraints, using gradient-based methods to solve the optimization problem. Their work provides theoretical guarantees, establishing a suitable approximation factor given the problem's constraints. 
More recently, a novel architectural approach has been proposed in \cite{de_neural_nodate}, which not only preserves submodularity but also extends its applicability to a more generalized form, accommodating $\alpha$-submodular functions. This signifies a notable advancement in the landscape of neural network-based modeling of submodular functions, expanding the scope of potential applications and providing a platform for exploring more nuanced and versatile representations within this domain. 
There are other works that attempt to represent monotone set functions using neural networks. For example, in \cite{weissteiner2021monotone}, the authors introduce a neural network architecture that represents all monotone set functions and explore their use in designing auctions. 

There are also works that approach the problem of learning (submodular) set functions from different angles: The authors in \cite{balcan_submodular_2018} investigate submodular functions from a learning theory perspective, developing algorithms for learning these functions and establishing lower bounds on their learnability. Moreover, the authors in \cite{feldman_learning_2014} attempt to approximate and learn coverage functions in polynomial time. One of the recent approaches to train and learn set functions using modern machine learning is  presented in \cite{NIPS2017_f22e4747}, which defines the properties of permutation invariant functions and presents a set of functions encompassing all permutation invariant objective functions. 

\section{Background}
\label{sec:back}
For a ground set $S$, any function $f: 2^S \rightarrow \mathbb{R}$ is referred to as a set function. Below, we present the definitions of essential concepts needed for the remainder of the discussion. 

\begin{definition}{(Monotone Set Function)}
	A set function $f: 2^S \rightarrow \mathbb{R}^+$ is called a monotone set function if for any $A \subseteq B \subseteq S$ we have,
\begin{equation}
f(A) \leq f(B).
\end{equation}

\end{definition} 

\begin{definition}{(Normalized Set Function)}
	A set function $f: 2^S \rightarrow \mathbb{R}^+$ is called a normalized set function if we have,
\begin{equation}
f(\emptyset) = 0.
\end{equation}
\end{definition} 

\begin{definition}{(Modular Function)}
	A function $m: 2^S \rightarrow \mathbb{R}^+$ is called a modular function if we have,
\begin{equation}
	\forall A \subseteq S: m(A) = \sum_{a\in A} m(a).
\end{equation}
\end{definition} 

Now, we can formally define submodular functions as follows.
\begin{definition}{(Submodular Function)}
A function $f: 2^S \rightarrow \mathbb{R}$, where $S$ is a finite set, is \textit{submodular} if for any $A \subseteq B \subseteq S$ and $v \notin B$, we have,
\begin{equation}
	f(A \cup v) - f(A) \geq f(B\cup v) - f(B).
\end{equation}
\end{definition}

In the remainder of the paper, without loss of generality, we will focus on normalized monotone set/submodular functions. If the function is not normalized, we can simply subtract the value of $f(\emptyset)$ from the function so as to make it normalized. Note that this transformation also maintains the submodularity of the function. 

\begin{definition}{(Sum of Concave Composed with Modular Functions \cite{bilmes_deep_2017})}
	Assume a finite set $S$ (\textit{nodes} or \textit{input features}) with cardinality $n$. Given a set of $m_1, m_2, \ldots, m_k$ ($m_i: 2^S \rightarrow \mathbb{R}^+$) modular functions and $\phi_1, \phi_2, \ldots, \phi_k$ ($\phi_i: \mathbb{R}^+ \rightarrow \mathbb{R}^+$) being their corresponding \textbf{non-negative}, non-decreasing, normalized (i.e., $\phi_i(0) = 0, \forall i$), concave functions, and an arbitrary modular function $m_{\pm}: 2^S \rightarrow \mathbb{R}^+$, the \textit{SCMM}, $g: 2^S \rightarrow \mathbb{R}^+$, derived by these functions is defined as,
\begin{equation}
\label{eq:scmm}
	g(A) = \sum_{i=1}^k \phi_i(m_i(A)) + m_{\pm}(A).
\end{equation}
\end{definition}

Based upon recent developments, it has been discovered that the aforementioned functions given in Equation~\ref{eq:scmm} exhibit inherent submodular characteristics \cite{bilmes_deep_2017}. 
Looking ahead, we can view these functions as a single-layer neural networks equipped with nonlinear activation functions that exhibit concavity properties, similar to a linear mixture of inputs followed by a stepwise activation function such as ReLU (Rectified Linear Unit) after computing the overall output values.

By employing this insight, we can expand the horizons of submodular functions by leveraging them across multiple tiers. We now introduce a pivotal concept that facilitates the utilization of advanced artificial intelligence tools, specifically deep learning.

\begin{figure}
	\centering
	\includegraphics[width=.5\textwidth]{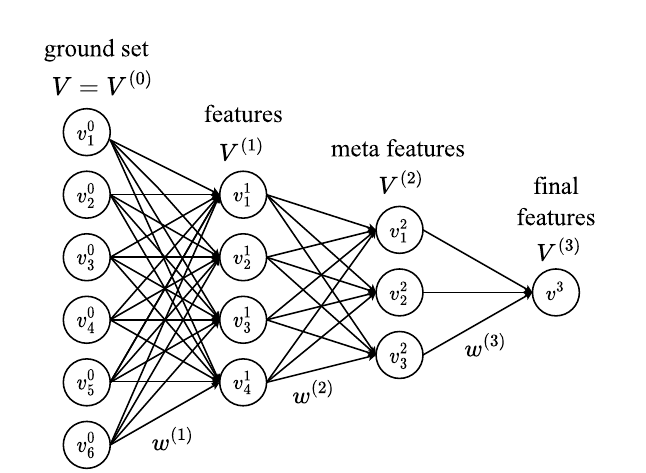}
	\caption{An example of deep submodular function \cite{bilmes_deep_2017}.}
    \label{fig:dsf-ex}
\end{figure}

\begin{definition}{(Deep Submodular Function \cite{bilmes_deep_2017})}
	Assume we intend to define $L$ (i.e., depth of the network) layer DSF. If $L = 1$ we use an SCMM (without last summation). Therefore, in this case we have a single-layer $n\times k$ network (with \textit{non-negative} weights) with submodular outputs. For $L = l > 1$ we first assume the output of $l-1$  layers as a given $n \times k$ DSF. Let’s denote the output nodes of the given DSF by $B = \{ \varphi_1, \varphi_2, \ldots, \varphi_k \}$. Now we want to add one layer at the end of the given DSF. We append a $k \times m$ fully connected layer with \textit{non-negative} weights and corresponding normalized non-decreasing concave functions $\phi_1, \phi_2, \ldots, \phi_m$ for each newly added node. Then we define the outputs of the network as  $C = \{ \psi_1, \psi_2, \ldots, \psi_m \}$
\begin{equation}
	\forall v \in C: \psi_v = \phi_v \left(\sum_{u \in [k]} w_{uv} \varphi_u\right) + b_v,
\end{equation}
and $b_v \in \mathbb{R}^+$ is a bias parameter of the node. In this scheme, we have new layer added to the DSF. Therefore, we introduced a $n\times m$ DSF with $L = l$ layers. As an example, a 3-layer DSF is shown in Figure \ref{fig:dsf-ex}.
\end{definition} 


With this understanding, one can guarantee that the outcome of this architecture forms a submodular function, as it is shown in \cite{bilmes_deep_2017}.
An established finding regarding SCMMs states that any SCMM employing an arbitrary activation function can be represented as a two-layer SCMM with only the $\min$ activation function \cite{bilmes_deep_2017}.


Despite these findings, there exist notable limitations when it comes to DSFs. While DSFs possess a range of capabilities, they are unable to encompass all submodular functions. This implies that there will always be submodular functions that cannot be represented using any number of layers in DSFs \cite{bilmes_deep_2017}. To address this limitation, in Section \ref{sec:edsf}, we extend DSFs by adding a limited number of components in the network architecture to represent all submodular functions.

In the following, we point out certain fundamental concepts in combinatorial optimization that will be employed throughout the remainder of the paper. We commence with an exploration of polymatroids.

\begin{definition}{(Polymatroid)}
	Consider a finite set $S$ with $|S| = s$ and a submodular function $f: 2^S \rightarrow \mathbb{R}$ defined on $S$. A polymatroid corresponding to $f$, denoted by $\mathcal{P}_f$, is defined as,
	\begin{equation}
		\mathcal{P}_f = \left\{ \mathbf{x} \in \mathbb{R}^s: \forall A \subseteq S: \mathbf{x}(A) \leq f(A), \mathbf{x} \succeq 0 \right\},
	\end{equation}
	where $\mathbf{x}(A) \coloneq \sum_{i\in A} x_i$.
\end{definition}

The following lemma describes an important property of the polymatroids corresponding to submodular functions.

\begin{lemma}{(\cite{goemans_approximating_2009})}
\label{lm:poly}
	For a polymatroid $\mathcal{P}_f$ corresponding to a monotone submodular function $f$, we have
	\begin{equation}
		\forall A \subseteq S: f(A) = \sup_{\mathbf{x} \in \mathcal{P}_f} \mathbf{x}(A).
	\end{equation}
\end{lemma}

We also define the following important submodular problem which we need later in our experiments.

\begin{definition}{(Submodular Welfare Maximization)}
	For a collection of $n$ users with $v_1, v_2, \ldots, v_n :2^S \rightarrow \mathbb{R}$ as their (estimated) \textit{submodular} valuation functions on each subset of the finite set $S$ with $|S| = s$, the submodular welfare maximization problem aims to maximize the social welfare function, i.e., the sum of all valuation functions when we \textit{partitioned} set of items $S$ and assigned to the users, and is formally defined as
\begin{align}
\label{eq:soc-wel-max}
	\max_{S_i, i=1,\ldots n} & \quad \sum_{i=1}^n v_i(S_i), \nonumber \\
s.t. & \quad S_i \subseteq S, \\
& \quad \bigcup_{i=1}^n S_i = S, \quad \forall i, j: S_i \cap S_j = \emptyset \nonumber .
\end{align}
Moreover, for any partition $A = \{S_1, S_2, \ldots, S_n\}$ of the ground set $S$ to users, we define its efficiency as, 
\begin{equation}
    \text{Eff}(A) \coloneq \frac{\sum_{i} v_i(S_i)}{\sum_i v_i(S^*_i)}
\end{equation}
where $OPT \coloneq \{S^*_1, S^*_2, \ldots, S^*_n\}$ is the optimal partition.
\end{definition}

Furthermore, in our experimental results, we will employe Coverage Functions, defined as follows.
\begin{definition}{(Coverage Function)}\label{def:coverage_function}
     A function $c : 2^{[n]} \rightarrow \mathbb{R}^+$ is a coverage function, if there exists a universe $U$ with non-negative weights $w(u)$ for each $u \in U$ and subsets $A_1, A_2, \ldots, A_n$ of $U$ such that for any $B\subseteq[n]$ we have $c(B) = \sum_{u \in \cup_{i\in B} A_i} w(u)$. 
\end{definition}
Alternatively, coverage functions can be described as non-negative linear combinations of monotone disjunctions. There are a natural subclass of submodular functions and arise in a number of applications \cite{feldman_learning_2014}.

\section{Extended Deep Submodular Functions}
\label{sec:edsf}
\subsection{Architecture Definition}
Formally, we define Extended Deep Submodular Function as follows.
\begin{definition}
    A set function $h$ is an EDSF if it can be represented as the minimum of $r$ Deep Submodular Functions $f_1,f_2,\ldots,f_r$, where $r$ is an arbitary number, namely,
    \begin{equation}
        h(A) = \min \left\{f_1(A), f_2(A), \ldots, f_r(A) \right\}, \quad \forall A \subseteq S.
    \end{equation}
\end{definition}

In the rest of this section, we aim to demonstrate that we can represent any monotone submodular function $f$ using some $g \in \text{EDSFs}$. The idea behind the proof is to leverage the relationship between the minimum of submodular functions and the intersection of their polymatroids.
The main result of this paper is summarized in the following theorem.
\begin{theorem}
    Family of monotone (submodular) set functions is exactly equal to the family of Extended Deep Submodular functions (EDSFs).
    \label{thm:main}
\end{theorem}

In the following, we will go to prove this result step by step. First, we assume that a submodular function $f: 2^S \rightarrow \mathbb{R}$ is given. For convenience, we use these notations to simplify our discussion.
\begin{definition}
We define:
	\begin{enumerate}
		\item $c_A = f(A)$, and for all $j \in S$ we denote $c_j = f(\{j\})$.
		\item a polytope $L_A$ for all $A \subseteq S$ as
		\begin{equation}
		L_A = \left\{\mathbf{x} \in \mathbb{R}^n: \mathbf{x} \succeq 0, \mathbf{x}(A) \leq c_A, \forall j \notin A: x_j \leq c_j \right\}.
		\end{equation}
		\item for any $A \subseteq S$ and $B \subseteq S$ the submodular function
		\begin{equation}
			g_A(B) = \min \left\{\sum_{j\in A \cap B} w_j, c_A \right\} + \sum_{k\in B \setminus A} w_k, 
		\end{equation}
		where $w_j = c_A$ for any $j\in A$ and $w_j = c_j$ for any $j \notin A$, and $w_j$ represents the corresponding weight of some neural network.
	\end{enumerate}
\end{definition}

\begin{center}
	\begin{figure}[h]
	\centering
	\includegraphics[width=.5\textwidth]{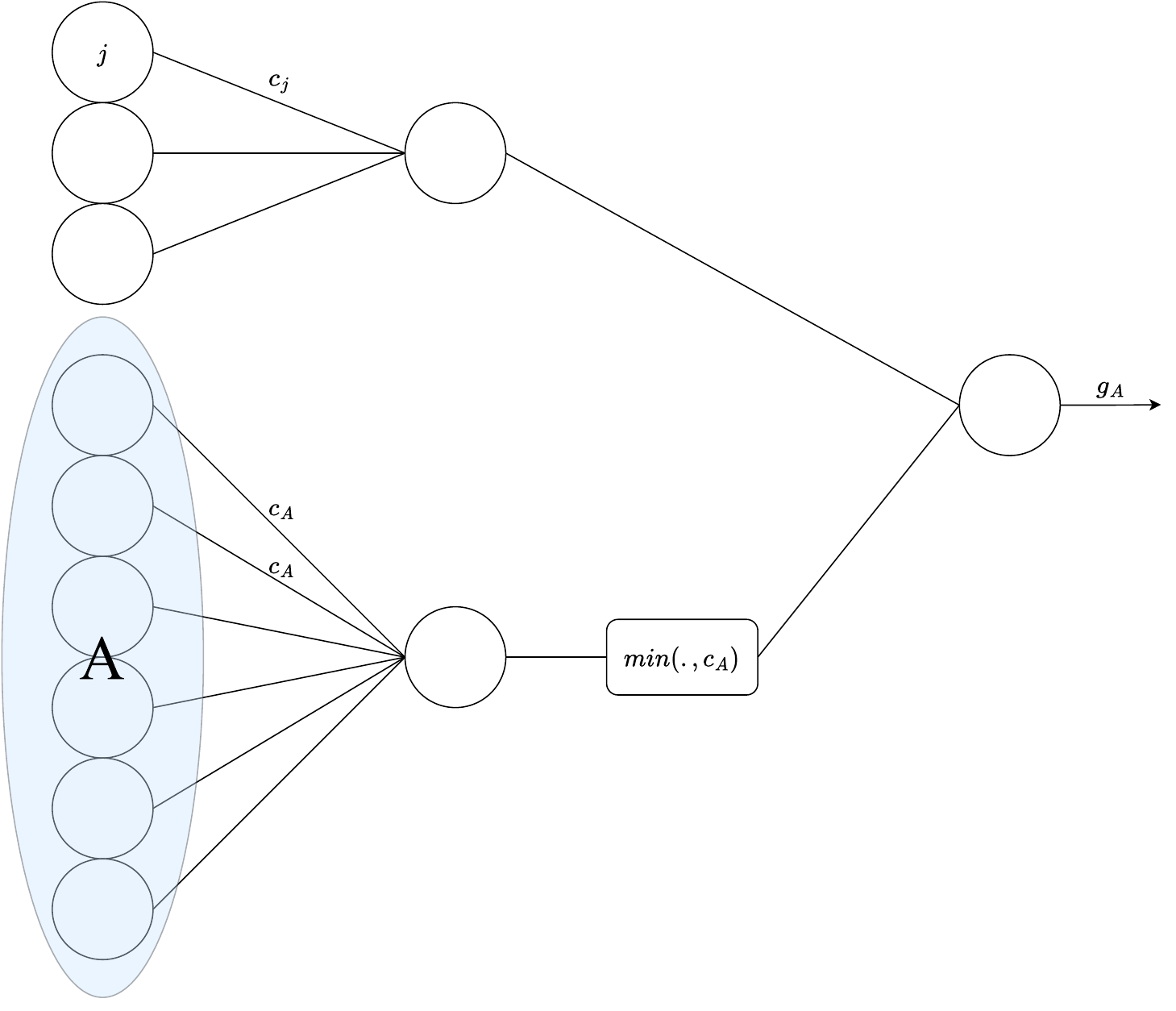}
	\caption{The simple architecture for the representation of $g_A$ as a DSF. The function $\min(\cdot,c_A)$ is the activation function of the lower node in the hidden layer. The upper hidden node has no activation function.}
\label{gad}
	\end{figure}
\end{center}

Using these definitions, in the first step, we will show that each $g_A$ can be represented using a simply two-layer DSF.
\begin{lemma}
	For any $A \subseteq S$, we can design a two-layer DSF that represents the $g_A$ function, therefore the $g_A$ function is a submodular function.
\end{lemma}
\proof To represent the function $g_A$ using a DSF, we have constructed a network architecture as illustrated in Figure \ref{gad}. This architecture comprises only two layers. The first layer contains two nodes, one of which employs the minimum function as its activation layer. The second layer consists of a single output node responsible for computing the sum of the inputs.

All edges connected to the lower node possess a weight of $c_A$. Conversely, each edge linked to the upper node bears a weight corresponding to the value $c_j$, specifically $f({j})$ according to the definition.

In this manner, we have introduced the network as a two-layer DSF
\begin{equation}
	g_A(B) = \min\left\{\sum_{j\in A \cap B} w_j, c_A \right\} + \sum_{k\in B \setminus A} w_k
\end{equation}
With this, we conclude our proof. \qed

We next introduce a useful lemma.
\begin{lemma}
\label{lm:ga}
	For any $A \subseteq S$, we have
	\begin{equation}
		\mathcal{P}_{g_A} = L_A.
	\end{equation}
\end{lemma}
\proof To show the equality of two aforementioned sets, we first show $\mathcal{P}_{g_A} \subseteq L_A$, then we show $L_A \subseteq \mathcal{P}_{g_A}$, which completes the proof.

For the first part, for any $\mathbf{x} \in \mathcal{P}_{g_A}$, we have for all $B \subseteq S$, $\mathbf{x}(B) \leq g_A(B)$, therefore, for any $B \subseteq A$, we have $\mathbf{x}(B) \leq c_A$, in fact in this scenario $g_A(B) = 0 \: \text{or} \: c_A$, that implies $\mathbf{x}(A) \leq c_A$. Furthermore, for any $j \notin A$ we have $x_j \leq f({j}) = c_j$. Therefore, each point in the polymatroid has the conditions to be in the $L_A$, that means, $\mathbf{x} \in L_A$.

For the reverse part, for any $\mathbf{x} \in L_A$, we have $\mathbf{x}(A) \leq c_A$, which implies that $\forall B \subseteq A, B \neq \emptyset$, $\mathbf{x}(B) \leq c_A = g_A(B)$. For any one-member subset of $S$, for example $B = \{j\}, j \notin A$ we have $x_j \leq c_j = g_A(B)$. Furthermore, for any arbitrary $B \subseteq S$, we could write $B = (A \cap B) \cup (B \setminus A)$. We have two cases,
\begin{align}
    1. & \ (A \cap B) = \emptyset \nonumber  \\
     & \implies \mathbf{x}(B) = \sum_{j\in B \setminus A}x_j \leq \sum_{j \in B \setminus A}c_j = g_A(B)  \nonumber \\
    2. & \ (A \cap B) \neq \emptyset \nonumber \\
      &\implies \mathbf{x}(B) = c_A + \sum_{j\in B \setminus A}x_j \leq c_A + \sum_{j \in B \setminus A}c_j = g_A(B). \nonumber
\end{align}

	It shows that for all $B \subseteq S$ we have: $\mathbf{x}(B) \leq g_A(B) \implies \mathbf{x} \in \mathcal{P}_{g_A}$. Therefore, the second part is now obvious, namely, $L_A \subseteq \mathcal{P}_{g_A}$. Hence,
\begin{equation}
	\begin{cases}
		\mathcal{P}_{g_A} \subseteq L_A \\
		L_A \subseteq \mathcal{P}_{g_A}
	\end{cases} \implies L_A = \mathcal{P}_{g_A},
\end{equation}
which completes the proof. \qed

For the next step, we introduce the next lemma.
\begin{lemma}
\label{lm:ispf}
For a given submodular function $f$, we have:
\begin{equation}
	\bigcap_{A \subseteq S} L_A = \mathcal{P}_f.
\end{equation}
\end{lemma}
\proof Similar to the proof of Lemma~\ref{lm:ga}, we first show that $\bigcap_{A \subseteq S} L_A \subseteq \mathcal{P}_f$, then we will prove $\mathcal{P}_f \subseteq \bigcap_{A \subseteq S} L_A$.

For the first part, for any $\mathbf{x} \in \bigcap_{A \subseteq S} L_A$ we have
\begin{equation}
	\forall A \subseteq S: \mathbf{x} \in  L_A \implies  \mathbf{x}(A) \leq c_A \implies \mathbf{x} \in \mathcal{P}_{f}.
\end{equation}
Therefore, we have shown that $ \bigcap_{A \subseteq S} L_A  \subseteq \mathcal{P}_f$. For the reverse direction, for any $\mathbf{x} \in \mathcal{P}_f$, we can write
\begin{equation}
	\forall A \subseteq S \implies \begin{cases}
		\mathbf{x}(A) \leq f(A) = c_A \\
		\forall j \notin A: x_j \leq f(\{j\}) = c_j
	\end{cases} \implies \mathbf{x} \in L_A.
\end{equation}
Therefore, we have shown that $\mathcal{P}_f \subseteq \bigcap_{A \subseteq S} L_A$. Combining these two parts completes our proof, namely, we have shown that $\bigcap_{A \subseteq S} L_A = \mathcal{P}_f$. \qed

For the next step, we need to present the following lemma.
\begin{lemma}
\label{lm:min}
	For a given set $f_1, f_2, \ldots, f_r$ of submodular functions, if the function $h = \min(f_1, f_2, \ldots, f_r)$ is submodular, for the polymatroid of the function $h$, we have
	\begin{equation}
		\mathcal{P}_h = \bigcap_{i=1, \ldots, r} \mathcal{P}_{f_i}.
	\end{equation}
\end{lemma}
\proof To prove the lemma we proceed as follows.
\begin{flalign*}
	\mathbf{x} \in \mathcal{P}_h &\iff \forall A \subseteq S: \mathbf{x}(A) \leq h(A) \\ 
	&\iff \forall A \subseteq S: \forall i: \mathbf{x}(A) \leq f_i(A) \\
	& \iff \forall i: \mathbf{x} \in \mathcal{P}_{f_i} \iff \mathbf{x} \in \bigcap_{i=1, \ldots, r} \mathcal{P}_{f_i}.
\end{flalign*}
\qed

In the next step, we introduce a new component to the DSF architecture to maintain the polymatroid corresponding to the output function to be the intersection of the polymatroids corresponding to the input functions. The proposed component is called \textit{Min-Component}, as defined below.
\begin{definition}{(Min-Component)}
	A Min-Component in the neural network, is a component (node) in which it  brings the minimum of the inputs as the output of the node.
\end{definition}

Incorporating this component into DSF, enhances its capability of representing submodular functions effectively. 


In general, note that the $\min$ operator does not maintain the submodularity of the input functions. However, during the proof of the Theorem \ref{thm:res}, we have used some techniques to assure that the output function would be submodular, as it can be seen in the following.

\begin{center}
	\begin{figure}[h]
	\centering
	\includegraphics[width=.4\textwidth]{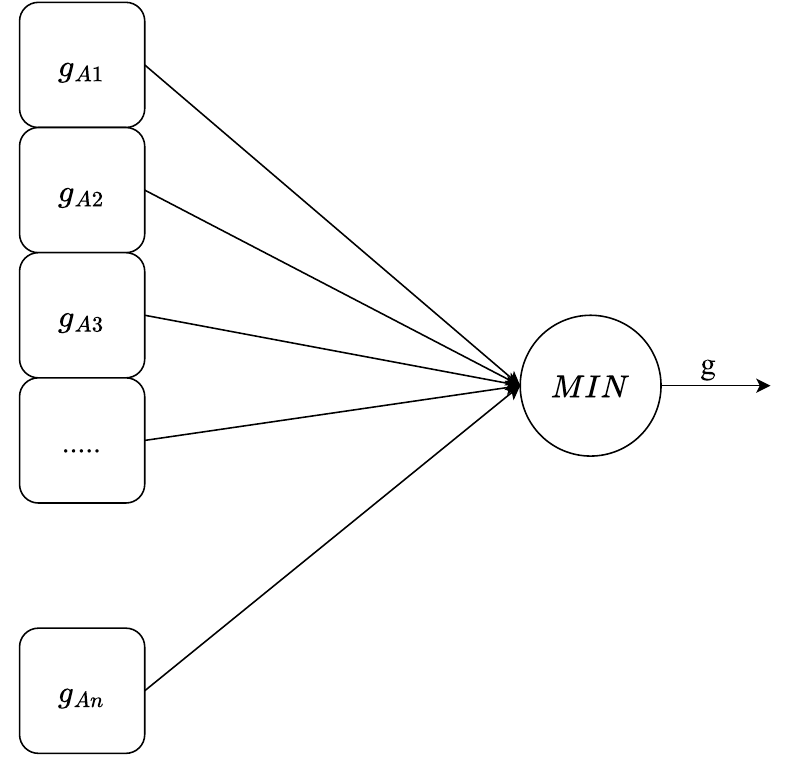}
	\caption{The architecture for the representation of the submodular function $f$.}
    \label{fig:arch}	
 \end{figure}
\end{center}

Moving towards the last stage, we present an architecture that is built upon the provided submodular function, as illustrated in Figure \ref{fig:arch}. In the initial layers, for any subset $A_i \subseteq S$, we have crafted the functions $g_{A_i}$ using the architecture outlined in Figure \ref{gad}. Subsequently, we applied a min-component to all of these constructed $g_{A_i}$ functions, resulting in a composite function denoted as $g$. Therefore, the corresponding EDSF is generated.

In the following theorem, we establish that the function $g$ can precisely represents the original submodular function $f$.

\begin{theorem}
	The function $g$ in Figure \ref{fig:arch} exactly represents the function $f$. In other words
	\begin{equation}
		\forall A \subseteq S: g(A) = f(A).
	\end{equation}
 \label{thm:res}
\end{theorem}
\proof Utilizing lemmas \ref{lm:ga}, \ref{lm:ispf}, and \ref{lm:min}, we can deduce that the polymatroid associated with the function $g$ is precisely identical to the polymatroid of the function $f$: $\mathcal{P}_g = \mathcal{P}_f$. However, it's worth noting that the function $g$ may belong to a broader category of functions than just submodular functions.

Based on lemma \ref{lm:poly}, we know that there exists some $\mathbf{x}^*$ such that $\mathbf{x}^* = \underset{\mathbf{x} \in \mathcal{P}_f}{\mathrm{argmax}}\, \mathbf{x}(A)$ and $\mathbf{x}^*(A) = f(A)$.
Now suppose that $f(A) > g(A)$. We can write
\begin{flalign}
\begin{cases}
	\mathbf{x}^* \in \mathcal{P} = \mathcal{P}_f = \mathcal{P}_g \\
	\mathbf{x}^*(A) = f(A)
\end{cases} 
 &\implies \mathbf{x}^*(A) = f(A) > g(A) \nonumber \\ &\implies \mathbf{x}^* \notin \mathcal{P}_g  \implies \mathbf{x}^* \notin \mathcal{P}_f.
\end{flalign}
That is a contradiction, so for any set $A \subseteq S$ we have $g(A) \geq f(A)$.

Furthermore, we can write
\begin{equation}
\begin{cases}
	g(A) = \min_{B \subseteq S} \left\{g_B(A) \right\} \\
	g_A(A) = c_A = f(A)
\end{cases} \implies g(A) \leq f(A).
\end{equation}
Combining these two results, it is clear that for any $A \subseteq S$ we have $g(A) = f(A)$.
\qed

Up to here, we have proved that all the monotone submodular functions can be represented by an EDSF function. In the remaining of this section, we will prove that this result also holds for the family of all monotone set functions.

\begin{lemma}
\label{lm:mono}
	For any EDSF function $f$, $f$ is monotone.
\end{lemma}
\proof The proof is straightforward, as all the weights used in construction of any EDSF is non-negative (all the constructing DSFs are monotone). All other functions that used in the construction of any EDSF such as $\min$ function are all monotone functions.
\qed

Next we prove the final result.
\begin{theorem}
    For any monotone set function $f$, there exists an EDSF $g$, such that $f = g$.
\end{theorem}
\proof Consider any monotone set function $f$. For any $B \subseteq S$ we define $g_B$ as follows
\begin{equation}
    g_B(A) = \min\left\{\sum_{j\in A \cap B} f(B), f(B) \right\} + \sum_{k\in A \setminus B} w^*
\end{equation}
where $w^* = f(S)$, which is the maximum value function $f$ can take (because $f$ is monotone). Now we define function $g$ as follows:
\begin{equation}
    g(A) = \min_{B \subseteq S} \left\{g_B(A) \right\}.
\end{equation}
For any $A \subseteq S$ we can see that $g_B(A) = f(A)$ if $B = A$. Now suppose $B \neq A$. If $A \setminus B \neq \emptyset$, $g_B(A) \geq f(S) \geq f(A)$, because of monotonicity. On the other hand if $A \setminus B = \emptyset$ we know that $A \subseteq B$. In this case we have $g_B(A) = f(B) \geq f(A)$
, because of monotonicity. Hence we can see that $g_A(A) \leq g_B(A)$ for any $B \subseteq S$. Therefore, $g(A) = f(A)$ for all $A \subseteq S$.
\qed

Based on the above results, we can conclude Theorem \ref{thm:main}.


\subsection{Concavity of EDSFs}
We can also show that any $g \in EDSFs$ is concave, if the input vector components are all non-negative real numbers, as stated in the following theorem.
\begin{theorem}
    Given $g \in EDSFs$, $g$ is a concave function with respect to the input vector, if the input vector components are all non-negative real numbers.
\end{theorem}
\proof Since all the DSF functions are concave in this setting \cite{bilmes_deep_2017}, and $g$ is the minimum of a number of DSFs and we know that minimum of concave functions are concave, we can conclude that $g$ is also concave. \qed

We can exploit the above-mentioned property to solve certain combinatorial optimization problems, such as the social welfare maximization problem, using gradient-based methods in an efficient manner. This provides a powerful tool to handle some combinatorial problems. Applications in this context are discussed in Section \ref{sec:experimental_resutls}.


\section{Experimental Results}\label{sec:experimental_resutls}
In the following section, we showcase a series of experiments aimed to demonstrating the positive outcomes and advantages derived from the application of Extended Deep Submodular Functions (EDSFs) in the modeling of submodular functions. Additionally, we highlight their efficacy in efficiently addressing and solving various combinatorial optimization problems. Through these experiments, we aim to provide a clear and comprehensive understanding of how EDSFs contribute to improved outcomes in the domain of submodular function modeling and the optimization of complex combinatorial scenarios.

\subsection{Learning Coverage Functions}
As outlined in Section \ref{sec:back}, coverage functions constitute a crucial and intricate subset of monotone submodular functions, posing challenges in accurate learning from their instances.

Our experimental findings indicate that, in contrast to Deep Submodular Functions (DSFs), Extended Deep Submodular Functions (EDSFs) shown to be effective in efficiently learning these complex functions, exhibiting much lower empirical generalization error, compared to DSFs.

To perform the experiment, from each coverage function (defined in Definition~\ref{def:coverage_function}), we generate a random dataset $\mathcal{D} = {(X_i, y_i)}_{i=1}^d$ where $X_i$ is a random subset from the ground set $S$, and $y_i$ is the value of the coverage function. In our experimental setup, we created three coverage functions with a universe size of 100 and 16 subsets. For each coverage function, we employed a Bernoulli random variable with probabilities of 0.1, 0.3, and 0.5, respectively, determining the likelihood of each item in the universe belonging to each subset independently. The weights function in the coverage function remains constant with a value of 1. We generated a dataset comprising 1024 samples, allocating 80\% for training and 20\% for testing.

We utilized quite similar architectures to train both Deep Submodular Functions (DSFs) and Extended Deep Submodular Functions (EDSFs) on the datasets, employing the $L1$-loss function. Subsequently, we calculated their respective empirical generalization errors on the corresponding test sets.

Notably, the training of DSFs on the above scenario proved to be unsuccessful, with the output remaining constant, likely due to the complexity of the coverage function. In contrast, EDSFs exhibited robust generalization with low error, showcasing their superior performance in this experimental context.

We report our final $L1$-loss, averaged on $100$ different experiments using DSFs and EDSFs in Table \ref{table:loss}.
\begin{table}[!h]
      \centering
        \begin{tabular}{ |c|c|c|c|c| } 
    \hline
    Coverage& Training & Test & Training & Test \\
     Prob.& Loss (EDSF) & Loss (EDSF) & Loss (DSF) & Loss (DSF) \\
    \hline
    0.1 & 1.0984 & 1.1127 & 25.0797 &  25.9907 \\ 
    0.3 & 1.0771 & 1.0913 & 13.3646 & 14.3648 \\ 
    0.5 & 0.6747 & 0.9852 & 6.2945 & 6.8395 \\
    \hline
\end{tabular}
    \caption{Training and Test $L1$-loss for EDSF and DSF setting.}
\label{table:loss}
\end{table}
The details of this experiment is elaborated in Appendix~\ref{apndx:coverage_function}.
 
Moreover, we conducted a series of experiments with different values of $r$, number of DSF used before using the $\min$ function, to show the effect of the size of the EDSF network on the generalization error of the training. In Table \ref{table:multiple-n}, we have shown the test loss for various size of $r$, using the same setting we discussed earlier.

\begin{table}[h]
\centering
\begin{tabular}{ |c|c| } 
    \hline
    r & Test Loss \\
    \hline
    1 &  82.1343  \\ 
    2 &  82.1319  \\ 
    4 & 81.9668  \\ 
    8 &  81.7255  \\ 
    16 &   3.3907  \\ 
    32 &   4.0554  \\ 
    64 &   3.1114  \\ 
    128 &   3.5768  \\ 
    256 &  16.1887  \\ 
    512 &   5.9880  \\  
    1024 &   8.9269  \\  
    \hline
\end{tabular}
\begin{center}
\caption{Experiments for various size of EDSF in learning coverage functions. Here, $r$ is the number of DSF functions used in constructing the EDSF.}
\label{table:multiple-n}
\end{center}
\end{table}

We also conducted some experiments of learning coverage function with universe size of 1000, probability of 0.2, and item numbers of 50 (much larger universe size and more items, compared to the previous experiment). Firstly, we kept the same architecture for both EDSFs and DSFs and ran 20 different experiments, which the average results are shown in Table \ref{table:edsf-test-loss-mean-std}.

\begin{table}[h]
\centering
\begin{tabular}{ |c|c|c| } 
    \hline
    Arch. & Test Loss Mean & Test Loss Std. \\
    \hline
    EDSF &  5.33770 & 1.3630 \\ 
    DSF &  82.58076 & 9.0195 \\ 
    \hline
\end{tabular}
\begin{center}
\caption{Experiments for comparison of EDSF and DSF for learning coverage functions with a large universe of size $1000$.}
\label{table:edsf-test-loss-mean-std}
\end{center}
\end{table}
For another set of experiments, we tried different architectures for DSFs, and the result of the training is also much worse than the above EDSF's result, as can be observed below.
\begin{enumerate}
    \item 5 hidden layers with 64 neurons each, alpha set to 95: test loss is 84.3058
    \item 4 hidden layers with 64 neurons each, alpha set to 125: test loss is 84.7727
    \item 5 hidden layers with respectively 32, 64, 128, 64, and 32 neurons, alpha set to 400: test loss is 81.4188
\end{enumerate}
The above experimental results demonstrate the superiority of EDSF compared to DSF in learning the coverage functions.

\subsection{Learning Cut Functions}
We conducted several experiments to learn modified graph cut functions (we choose modified version to maintain the monotonicity), which is a well-known submodular function. Firstly, we generated the random graph using Erdos-Renyi model (with probability 0.2 and 50 vertices), then, for any set of vertices $X$, we considered the function $f(X) = |\mathrm{cut}(X)| + \sum_{a\in X} \mathrm{deg}(a)$ which is a monotone submodular function. To learn the cut function, we have used the EDSF architecture with 64 DSFs as its backbone. The result is that for 12 trials, we have (mean, standard deviation) of test loss: (4.6309, 0.5857).
The additional figures are presented in Appendix~\ref{apndx:cut_function}. Our experiments also demonstrated that in this particular scenario, DSFs also perform reasonably in learning cut functions.

\subsection{Social Welfare Maximization}
In this experiment, we leveraged the previously discussed concave property inherent in EDSFs to address the widely recognized combinatorial optimization problem of maximizing social welfare, as delineated in Section \ref{sec:back}. In this specific context, each of the valuation functions, denoted as $v_1, v_2, \ldots, v_n$, is assumed to be an EDSF or a DSF, acquired through learning from samples collected from users. Our optimization problem, expressed in Equation \ref{eq:soc-wel-max}, can be reformulated as follows,
\begin{align}
	\max_{a} & \quad SW(a) \delequal \sum_{i=1}^n v_i(a_i), \nonumber \\
s.t. &  \quad \sum_{i=1}^n a_{ij} = 1, \text{for all} \quad j \in S \nonumber, \\
& \quad a_{ij} \in \{0,1\}.
\end{align}
where $a$ is a $n\times m$ matrix, such that $a_{ij} \in \{0,1\}$ meaning if item $j$ assigned to user $i$ and $a_i$ represents the $i$'th row of $a$, which is the items that were assigned to user $i$.

To solve this optimization problem, first we relax it to the following form,
\begin{align}
	\max_{a} & \quad SW(a) = \sum_{i=1}^n v_i(a_i), \nonumber \\
s.t. &  \quad \sum_{i=1}^n a_{ij} = 1, \text{for all} \quad j \in S \nonumber, \\
& \quad a_{ij} \in [0,1].
\end{align}
As we mentioned, each of the $v_i$'s are EDSF or DSF, so they are concave, hence, the relaxed problem is clearly a convex problem, e.g., maximizing a concave function with convex constraints, and can be solved using convex optimization techniques such as projected gradient ascent.
The projection step consists of $m$ distinct projections on probability simplex for each item. The algorithm pseudo-code is shown in Algorithm \ref{alg:grad-asc}.
\begin{algorithm}[tb]
   \caption{Gradient Ascent}
   \label{alg:grad-asc}
\begin{algorithmic}
   \STATE {\bfseries Input:} valuation functions $v_1, v_2, \ldots, v_n$, set of items $S$ ($|S|=s$), learning rate $\eta$
   \STATE Initialize $a = (0)_{ij}$
   \STATE Project each column of $a$ on the probability simplex
   \REPEAT
   \STATE Compute gradient of $SW$ function in the point $a$ $
   = g \in \partial SW$. 
   \STATE $a = a + \eta . g$
    \STATE Project each column of $a$ on the probability simplex
   \UNTIL{convergence}
   \FOR{$i$ in $S$}
   \STATE Select the user assigned to item $i$ by sampling from $i$'th column of $a$ (corresponding distribution for item $i$).
   \ENDFOR
\end{algorithmic}
\end{algorithm}

In this experiment, we initially presumed that the true submodular value functions were Coverage Functions. Subsequently, we trained both Extended Deep Submodular Functions (EDSFs) and Deep Submodular Functions (DSFs) using pre-defined architectures for each user valuation function. We then employed gradient ascent on the social welfare function to find semi-optimal item assignment.
\begin{table}[h]
\centering
\begin{tabular}{ |c|c|c|c| } 
    \hline
    Exp. & Predicted &Optimal&Efficiency \\
    No. & Social Welfare & Social Welfare & \% \\
    \hline
    1 & 124 & 127 & 97.6378 \\ 
    2 & 114 & 123 & 92.6829 \\ 
    3 & 84 & 129 & 65.1163 \\ 
    4 & 118 & 128 & 92.1875 \\ 
    5 & 114 & 127 & 89.7638 \\ 
    6 & 126 & 128 & 98.4375 \\ 
    7 & 110 & 122 & 90.1639 \\ 
    8 & 123 & 132 & 93.1818 \\ 
    9 & 123 & 128 & 96.0938 \\ 
    10 & 120 & 129 & 93.0233 \\  
    \hline
    Average & 115.6 & 127.3 & 90.8288 \\
    \hline
\end{tabular}
\begin{center}
\caption{Experiments when value functions are coverage functions, with 50 universe size, and 8 items, with probablities 0.1, 0.3, and 0.5 for three bidders respectively, and the learning model is EDSF.}
\label{table:CoveragewithEDSF}
\end{center}
\end{table}

\begin{table}[h!]
\centering
\begin{tabular}{ |c|c|c|c| } 
    \hline
    Exp.& Predicted &Optimal&Efficiency \\
    No. & Social Welfare & Social Welfare & \% \\
    \hline
    1 & 98 & 127 & 77.1654 \\ 
    2 & 98 & 121 & 80.9917\\ 
    3 & 63 & 130 & 48.4615\\ 
    4 & 50 & 124 & 40.3226\\ 
    5 & 94 & 120 & 78.3333\\ 
    6 & 92 & 131 & 70.2290\\ 
    7 & 92 & 126 & 73.0159\\ 
    8 & 99 & 129 & 76.7442\\ 
    9 & 94 & 127 & 74.0157\\ 
    10 & 60 & 125 & 48.0000\\  
    \hline
    Average & 84 & 126 & 66.7279 \\
    \hline
\end{tabular}
\begin{center}
\caption{Experiments when value functions are coverage functions, with 50 universe size, and 8 items, with probablities 0.1, 0.3, and 0.5 for three bidders respectively, and the learning model is DSF.}
\label{table:CoveragewithDSF}
\end{center}
\end{table}
Furthermore, we conducted a new experiment to compare the performance of vanilla nerual networks versus EDSFs. The results are shown in Table \ref{table:CoveragewithVanilla}.
We can see that the mean efficiency is about 77\%, which is much lower than the EDSF efficiency (90\%).
Also, we have conducted experiments for universe size 500 and 8 sets (items) of the coverage function. The Table \ref{table:Coveragewith500} shows the results.
Finally, we conducted experiments for universe size 1000 and 8 sets (items) for the coverage function. The Table \ref{table:Coveragewith1000} shows the results about efficiency of the algorithms.
\begin{table}[h!]
\centering
\begin{tabular}{ |c|c|c|c| } 
    \hline
    Exp.& Predicted &Optimal&Efficiency \\
    No. & Social Welfare & Social Welfare & \% \\
    \hline
    1 & 56.0 & 98.0	& 57.14 \\ 
    2 & 79.0 & 101.0 & 78.28\\ 
    3 & 89.0 & 93.0	& 95.70\\ 
    4 & 81.0 & 94.0	& 86.17\\ 
    5 & 70.0 & 99.0	& 70.71\\ 
    6 & 52.0 & 94.0	& 55.32\\ 
    7 & 84.0 & 100.0 & 84.00\\ 
    8 & 62.0 & 101.0 & 61.39\\ 
    9 & 88.0 & 99.0	& 88.89\\ 
    10 & 81.0 & 96.0 & 84.37\\ 
    11 & 77.0 & 93.0 & 82.80\\ 
    12 & 92.0 & 108.0 & 85.19\\ 
    13 & 77.0 & 104.0 & 74.04\\ 
    14 & 85.0 & 99.0 & 85.86\\ 
    15 & 69.0 & 94.0 & 73.40\\  
    \hline
\end{tabular}
\begin{center}
\caption{Experiments when value functions are coverage functions, with 50 universe size, and 8 items, with probablities 0.1, 0.3, and 0.5 for three bidders respectively, and the learning model is vanilla neural network.}
\label{table:CoveragewithVanilla}
\end{center}
\end{table}

\begin{table}[h!]
\centering
\begin{tabular}{ |c|c|c|c|c|c| } 
    \hline
    Exp.& EDSF Predicted & DSF Predicted & Optimal& EDSF & DSF \\
    No. & Social Welfare & Social Welfare & Social Welfare & Eff. \% & Eff. \% \\
    \hline
    1 & 892.0 & 653.0 & 902.0  & 98.89 & 72.39  \\ 
    2 & 907.0 & 634.0 & 917.0  & 98.91 & 69.14 \\ 
    3 & 887.0 & 636.0 & 892.0 & 99.44 & 71.30 \\ 
    4 & 882.0 & 500.0 & 900.0 & 98.00 & 55.56 \\ 
    5 & 909.0 & 670.0 & 913.0 & 99.56 & 73.38 \\ 
    \hline
    Average & 894.0	&  618.0 & 904.0 & 98.96 & 68.35 \\
    \hline
\end{tabular}
\begin{center}
\caption{Experiments comparing EDSF and DSF efficiency in the maximizing social welfare problem, with coverage function as value function, 500 universe size, with probabilities 0.1, 0.3, and 0.5 for three bidders, respectively.}
\label{table:Coveragewith500}
\end{center}
\end{table}

\begin{table}[h!]
\centering
\begin{tabular}{ |c|c|c|c|c|c| } 
    \hline
    Exp.& EDSF Predicted & DSF Predicted & Optimal& EDSF & DSF \\
    No. & Social Welfare & Social Welfare & Social Welfare & Eff. \% & Eff. \% \\
    \hline
    1 & 1737.0 & 1527.0 & 1782.0  & 97.47 & 85.69  \\ 
    2 & 1742.0 & 1730.0 & 1789.0  & 97.37 & 96.70 \\ 
    3 & 1744.0 & 1547.0 & 1778.0 & 98.09 & 87.01 \\ 
    4 & 1784.0 & 1739.0 & 1790.0 & 99.66 & 97.15 \\ 
    5 & 1755.0 & 1699.0 & 1776.0 & 98.82 & 95.66 \\ 
    \hline
    Average & 1752.4 &  1648.4 & 1783.0 & 98.2 & 92.60 \\
    \hline
\end{tabular}
\begin{center}
\caption{Experiments comparing EDSF and DSF efficiency in the maximizing social welfare problem, with coverage function as value function, 1000 universe size, with probabilities 0.1, 0.3, and 0.5 for three bidders, respectively.}
\label{table:Coveragewith1000}
\end{center}
\end{table}

Observing the results, it becomes apparent that the utilization of Extended Deep Submodular Functions (EDSFs) in the context of the social welfare maximization problem yields significantly higher efficiency when compared to Deep Submodular Functions (DSFs) and also vanilla neural networks. This stark difference in efficiency underscores the potential advantages and superior performance that our proposed framework, leveraging EDSFs, can bring to the modeling of user valuations within the realm of this NP-hard combinatorial optimization problem. 

\section{Discussion}
In this paper, we introduced an architecture to represent all monotone set/submodular functions using neural networks. Here are some points about the proposed architecture that are worth-noting. The most important shortcoming of this architecture is the exponential size of the network to represent all monotone set/submodular functions. However, we would like to note that the assumption on the exponentiality of the number of DSFs is just used as a sufficient condition for our proofs. In our experiments, we observed that using much less number of DSFs than exponential order is enough to attain good generalization in practice. For example, in the problem of learning coverage functions, we used 64 DSFs to represent a function with 16 and 50 items as input (note that $64 \ll 2^{16}$ and $64 \ll 2^{50}$). We have also used 64 DSFs in learning cut functions of graphs with 50 vertices ($64 \ll 2^{50}$).


On the other hand, there exist some approximation algorithm to find the near-optimal social welfare in the problem setting mentioned above, like the one proposed in \cite{vondrak_optimal_2008}, namely, Randomized or Continuous Greedy.
We have conducted multiple experiments in order to compare the Continuous Greedy and Gradient Ascent algorithm as follows. We performed 20 trials with 16 items, in the setting of 3 participants with coverage functions as valuation functions that have a universe size of 1000, and various probabilities. The mean of (Gradient Ascent, Continuous Greedy) over multiple trials to find maximum social welfare is (2333.45, 2332.2).
As can be seen from the results, the ``average'' performance of both methods (Gradient Ascent and Continuous Greedy) is very similar, leading us to conclude that there may be a shared intuition underlying both algorithms. The intuition behind the continuous greedy algorithm is that the current distribution will shift towards the most increasing direction to find the optimal distribution for sampling. This concept is reminiscent of moving in the direction of the gradient in concave functions to maximize them. Overall, it appears that the Gradient Ascent algorithm closely resembles the continuous dual of the continuous greedy algorithm. Exploring the theoretical connection between Gradient Ascent and Continuous Greedy can be an interesting direction for future work.

However, in the following, we would like to mention that the continuous greedy algorithm has two major limitations.
\begin{enumerate}
    \item In every step, the Continuous Greedy (CG) algorithm needs to sample a large number of function calculations ($O((mn)^5)$) to determine the expected marginal increment. Moreover, it needs at least $O((mn)^2)$ steps to obtain the result. This imposes a significant computational burden on CG compared to Gradient Ascent (GA), where the gradient is calculated with much less computational complexity, since the computation of gradient is $O(1)$. This fact limits the scalability of the Continuous Greedy algorithm. we also conducted some experiments for 100 items and the GA algorithm is much faster than CG algorithm in practice.
    \item Furthermore, the proposed approach is applicable to both monotone submodular and non-submodular functions, while CG is only applicable to submodular functions.
\end{enumerate}

\section{Conclusions}
In this research, we introduce a novel concept called Extended Deep Submodular Functions (EDSFs), building upon the foundation of Deep Submodular Functions (DSFs). DSFs, a subset of monotone submodular functions, provide a structured framework for representing specific types (a subset) of submodular functions. However, the scope of DSFs is limited to a strict subset of monotone submodular functions. EDSFs, on the other hand, serve as a natural extension, expanding the family of DSFs to encompass all monotone set/submodular functions.
Our proofs are rooted in the properties of polymatroids, offering insights into the relationship between polymatroids and submodular functions. Additionally, we highlight the concave nature of EDSFs, a characteristic that is proved to be valuable in addressing and efficiently solving various combinatorial optimization problems.
To validate the efficacy of EDSFs, we conducted experiments in three distinct settings, namely, learning coverage functions, learning modified cut functions, and maximizing social welfare. The results consistently demonstrated the superior performance of EDSFs compared to DSFs in these experimental scenarios.
In conclusion, our findings suggest that EDSFs provide a more comprehensive and effective solution for modeling monotone set/submodular functions using neural networks. The extended scope and enhanced performance make EDSFs a promising direction for further exploration in various machine learning and optimization domains.

\section*{Broader Impact}
This paper presents work whose goal is to advance the field of Machine Learning. There aren't many straight potential societal consequences of our work, none which we feel must be specifically highlighted here.

\bibliography{tmlr}

\begin{thebibliography}{15}
\providecommand{\natexlab}[1]{#1}
\providecommand{\url}[1]{\texttt{#1}}
\expandafter\ifx\csname urlstyle\endcsname\relax
  \providecommand{\doi}[1]{doi: #1}\else
  \providecommand{\doi}{doi: \begingroup \urlstyle{rm}\Url}\fi

\bibitem[Bai et~al.(2018)Bai, Stafford~Noble, and Bilmes]{bai_submodular_2018}
Wenruo Bai, William Stafford~Noble, and Jeff~A Bilmes.
\newblock Submodular {Maximization} via {Gradient} {Ascent}: {The} {Case} of
  {Deep} {Submodular} {Functions}.
\newblock In \emph{Advances in {Neural} {Information} {Processing} {Systems}},
  volume~31. Curran Associates, Inc., 2018.
\newblock URL
  \url{https://proceedings.neurips.cc/paper_files/paper/2018/hash/b43a6403c17870707ca3c44984a2da22-Abstract.html}.

\bibitem[Balcan \& Harvey(2018)Balcan and Harvey]{balcan_submodular_2018}
Maria-Florina Balcan and Nicholas J.~A. Harvey.
\newblock Submodular {Functions}: {Learnability}, {Structure}, and
  {Optimization}.
\newblock \emph{SIAM Journal on Computing}, 47\penalty0 (3):\penalty0 703--754,
  January 2018.
\newblock ISSN 0097-5397, 1095-7111.
\newblock \doi{10.1137/120888909}.
\newblock URL \url{https://epubs.siam.org/doi/10.1137/120888909}.

\bibitem[Bilmes \& Bai(2017)Bilmes and Bai]{bilmes_deep_2017}
Jeffrey Bilmes and Wenruo Bai.
\newblock Deep {Submodular} {Functions}, January 2017.
\newblock URL \url{http://arxiv.org/abs/1701.08939}.
\newblock arXiv:1701.08939 [cs].

\bibitem[De \& Chakrabarti(2022)De and Chakrabarti]{de_neural_nodate}
Abir De and Soumen Chakrabarti.
\newblock Neural {Estimation} of {Submodular} {Functions} with {Applications}
  to {Differentiable} {Subset} {Selection}.
\newblock 2022.

\bibitem[Dolhansky \& Bilmes(2016)Dolhansky and Bilmes]{dolhansky2016deep}
Brian~W Dolhansky and Jeff~A Bilmes.
\newblock Deep submodular functions: Definitions and learning.
\newblock \emph{Advances in Neural Information Processing Systems}, 29, 2016.

\bibitem[Feldman \& Kothari(2014)Feldman and Kothari]{feldman_learning_2014}
Vitaly Feldman and Pravesh Kothari.
\newblock Learning coverage functions and private release of marginals.
\newblock In \emph{Conference on {Learning} {Theory}}, pp.\  679--702. PMLR,
  2014.
\newblock URL \url{https://proceedings.mlr.press/v35/feldman14a.html}.

\bibitem[Gillenwater et~al.(2012)Gillenwater, Kulesza, and
  Taskar]{gillenwater2012near}
Jennifer Gillenwater, Alex Kulesza, and Ben Taskar.
\newblock Near-optimal map inference for determinantal point processes.
\newblock \emph{Advances in Neural Information Processing Systems}, 25, 2012.

\bibitem[Goemans et~al.(2009)Goemans, Harvey, Iwata, and
  Mirrokni]{goemans_approximating_2009}
Michel~X. Goemans, Nicholas J.~A. Harvey, Satoru Iwata, and Vahab Mirrokni.
\newblock Approximating {Submodular} {Functions} {Everywhere}.
\newblock In \emph{Proceedings of the {Twentieth} {Annual} {ACM}-{SIAM}
  {Symposium} on {Discrete} {Algorithms}}, pp.\  535--544. Society for
  Industrial and Applied Mathematics, January 2009.
\newblock ISBN 978-0-89871-680-1 978-1-61197-306-8.
\newblock \doi{10.1137/1.9781611973068.59}.
\newblock URL \url{https://epubs.siam.org/doi/10.1137/1.9781611973068.59}.

\bibitem[Kempe et~al.(2003)Kempe, Kleinberg, and Tardos]{kempe2003maximizing}
David Kempe, Jon Kleinberg, and {\'E}va Tardos.
\newblock Maximizing the spread of influence through a social network.
\newblock In \emph{Proceedings of the ninth ACM SIGKDD international conference
  on Knowledge discovery and data mining}, pp.\  137--146, 2003.

\bibitem[Kimball et~al.(2024)Kimball, Reck, Zhang, Ohtake, and
  Tsutsui]{kimball2024diminishing}
Miles~S Kimball, Daniel Reck, Fudong Zhang, Fumio Ohtake, and Yoshiro Tsutsui.
\newblock Diminishing marginal utility revisited.
\newblock Technical report, National Bureau of Economic Research, 2024.

\bibitem[McLaughlin \& Chernew(2001)McLaughlin and
  Chernew]{mclaughlin2001health}
CG~McLaughlin and ME~Chernew.
\newblock Health insurance: economic and risk aspects.
\newblock 2001.

\bibitem[Narasimhan et~al.(2005)Narasimhan, Jojic, and Bilmes]{narasimhan2005q}
Mukund Narasimhan, Nebojsa Jojic, and Jeff~A Bilmes.
\newblock Q-clustering.
\newblock \emph{Advances in Neural Information Processing Systems}, 18, 2005.

\bibitem[Vondrak(2008)]{vondrak_optimal_2008}
Jan Vondrak.
\newblock Optimal approximation for the submodular welfare problem in the value
  oracle model.
\newblock In \emph{Proceedings of the fortieth annual {ACM} symposium on
  {Theory} of computing}, pp.\  67--74, Victoria British Columbia Canada, May
  2008. ACM.
\newblock ISBN 978-1-60558-047-0.
\newblock \doi{10.1145/1374376.1374389}.
\newblock URL \url{https://dl.acm.org/doi/10.1145/1374376.1374389}.

\bibitem[Weissteiner et~al.(2021)Weissteiner, Heiss, Siems, and
  Seuken]{weissteiner2021monotone}
Jakob Weissteiner, Jakob Heiss, Julien Siems, and Sven Seuken.
\newblock Monotone-value neural networks: Exploiting preference monotonicity in
  combinatorial assignment.
\newblock \emph{arXiv preprint arXiv:2109.15117}, 2021.

\bibitem[Zaheer et~al.(2017)Zaheer, Kottur, Ravanbakhsh, Poczos, Salakhutdinov,
  and Smola]{NIPS2017_f22e4747}
Manzil Zaheer, Satwik Kottur, Siamak Ravanbakhsh, Barnabas Poczos, Russ~R
  Salakhutdinov, and Alexander~J Smola.
\newblock Deep sets.
\newblock In I.~Guyon, U.~Von Luxburg, S.~Bengio, H.~Wallach, R.~Fergus,
  S.~Vishwanathan, and R.~Garnett (eds.), \emph{Advances in Neural Information
  Processing Systems}, volume~30. Curran Associates, Inc., 2017.
\newblock URL
  \url{https://proceedings.neurips.cc/paper_files/paper/2017/file/f22e4747da1aa27e363d86d40ff442fe-Paper.pdf}.

\end{thebibliography}
\bibliographystyle{tmlr}

\appendix
\section{Details of Experiments on Learning Coverage Functions}\label{apndx:coverage_function}
In the following experiments, we trained EDSFs and DSFs with similar architectures, except the min-component at the end of the EDSFs, consists of three fully-connected layers with 64 neurons stacked with $\phi(x) = \min(\alpha, x)$, i.e., minimum linear unit, with $\alpha=95$ as an activation function.

\begin{figure}[ht]
    \centering
    \includegraphics[width=.4\textwidth]{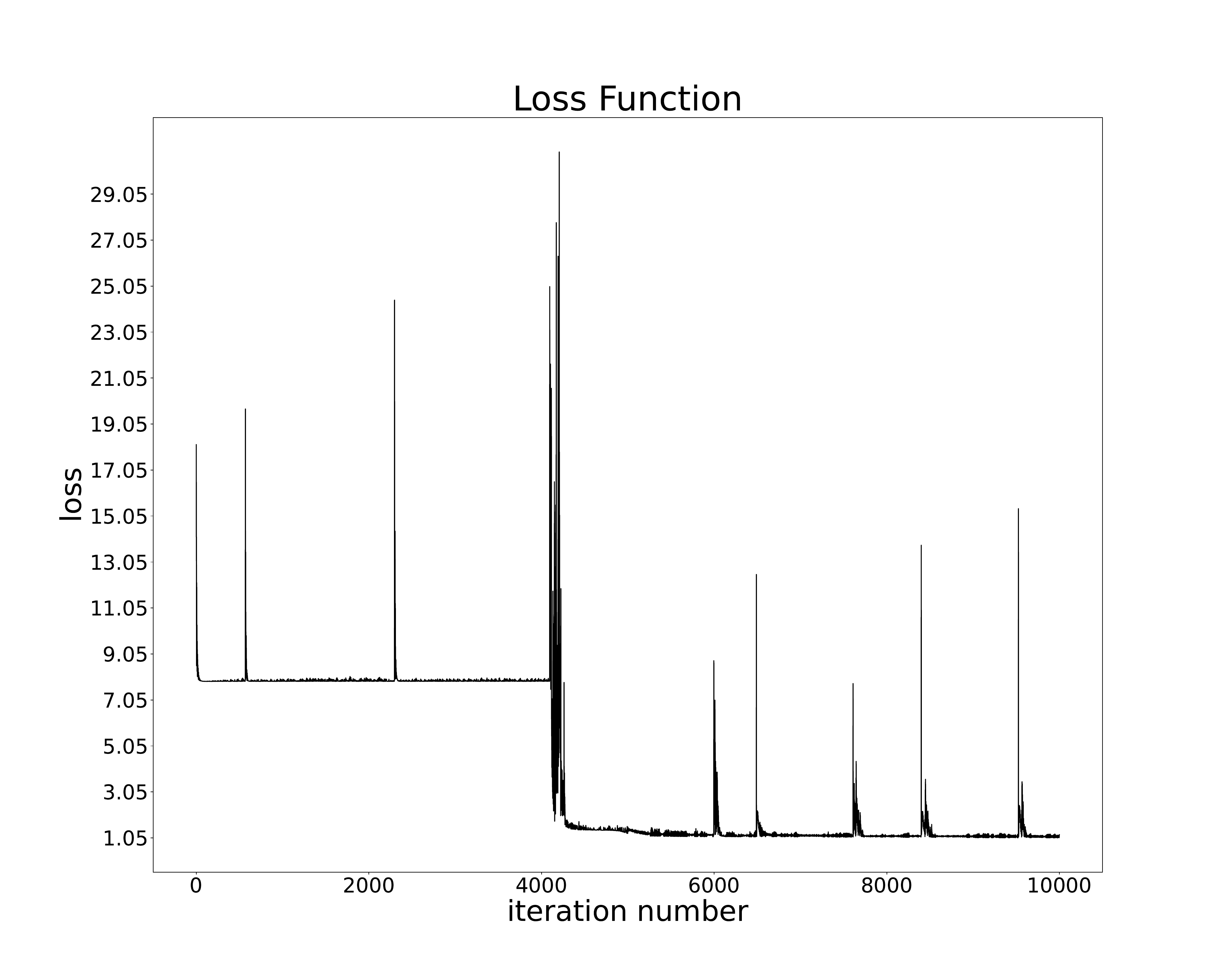}
    \includegraphics[width=.5\textwidth]{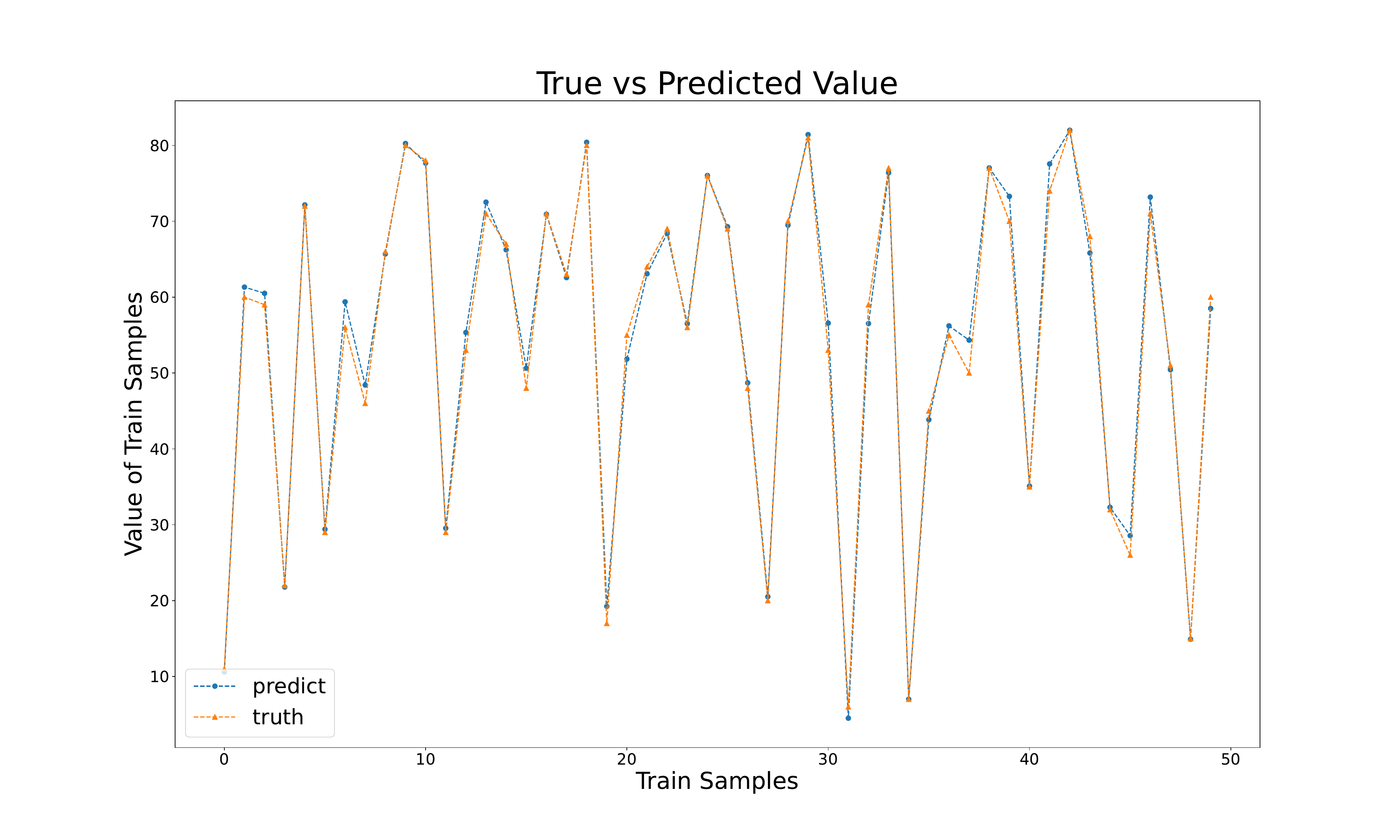}
    \includegraphics[width=.5\textwidth]{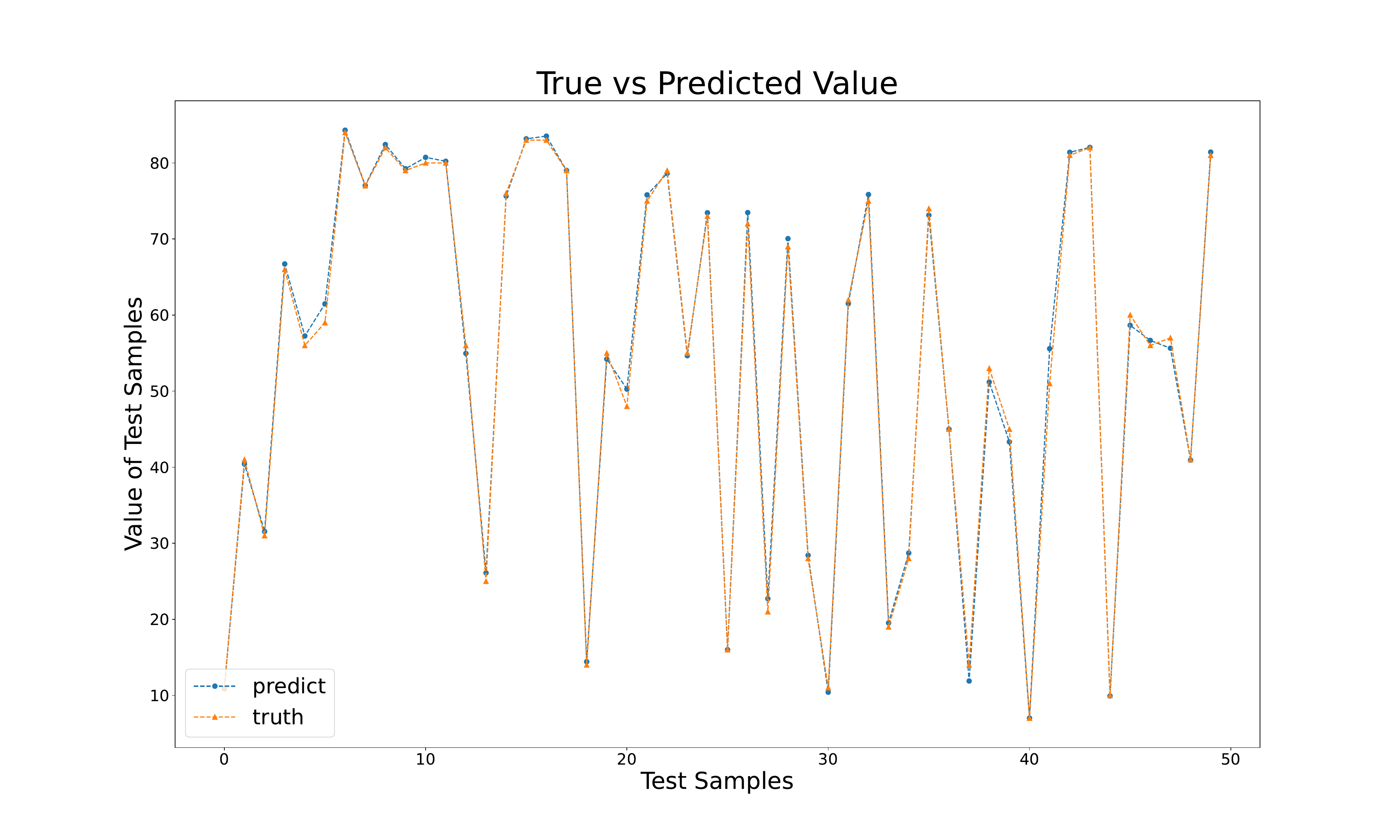}
    \caption{Learning coverage function with coverage probability 0.1, using the EDSF architecture, showing Training loss, Truth vs. Predicted values for train and test samples}
    \label{fig:edsf-low}
\end{figure}

\begin{figure}[ht]
    \centering
    \includegraphics[width=.6\textwidth]{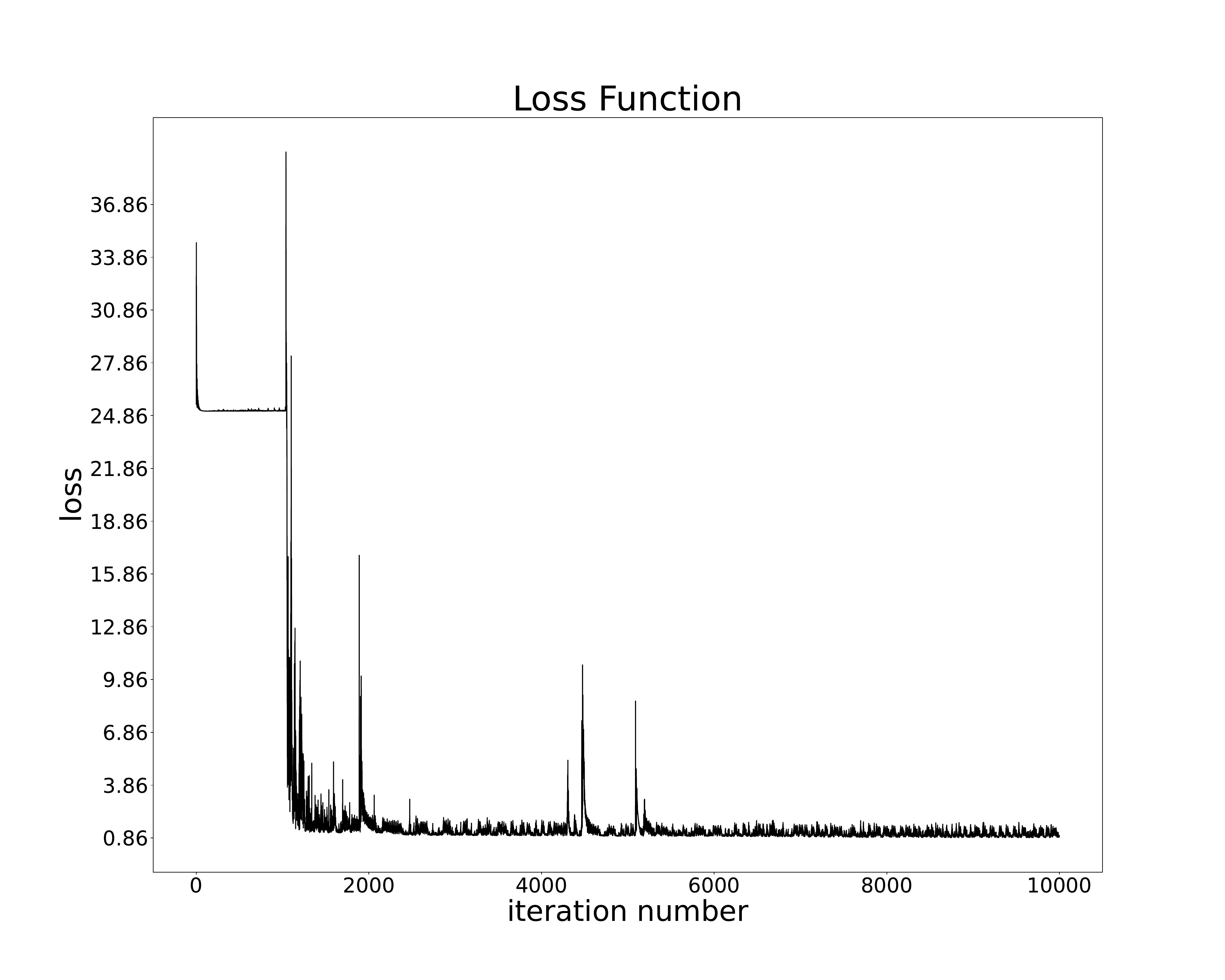}
    \includegraphics[width=.6\textwidth]{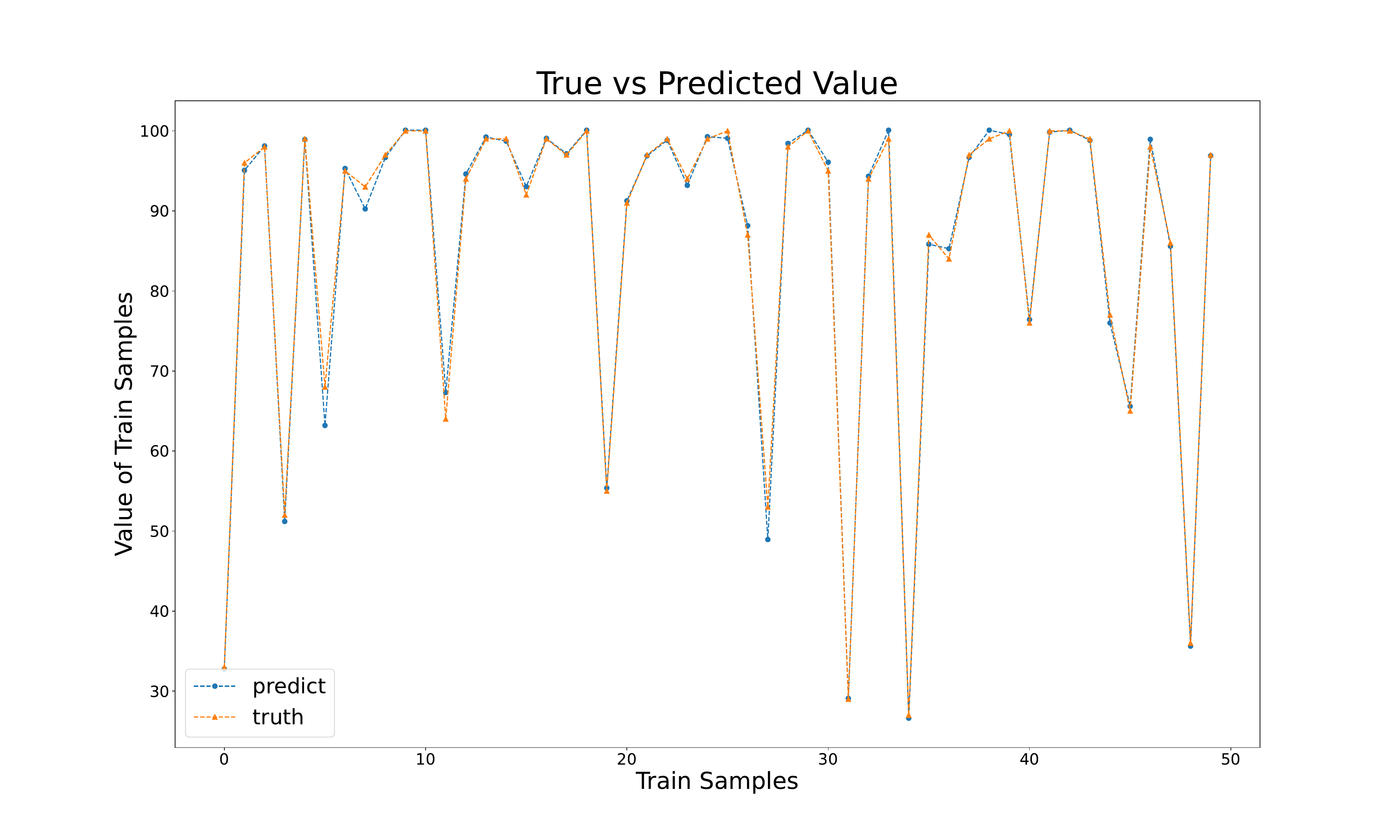}
    \includegraphics[width=.6\textwidth]{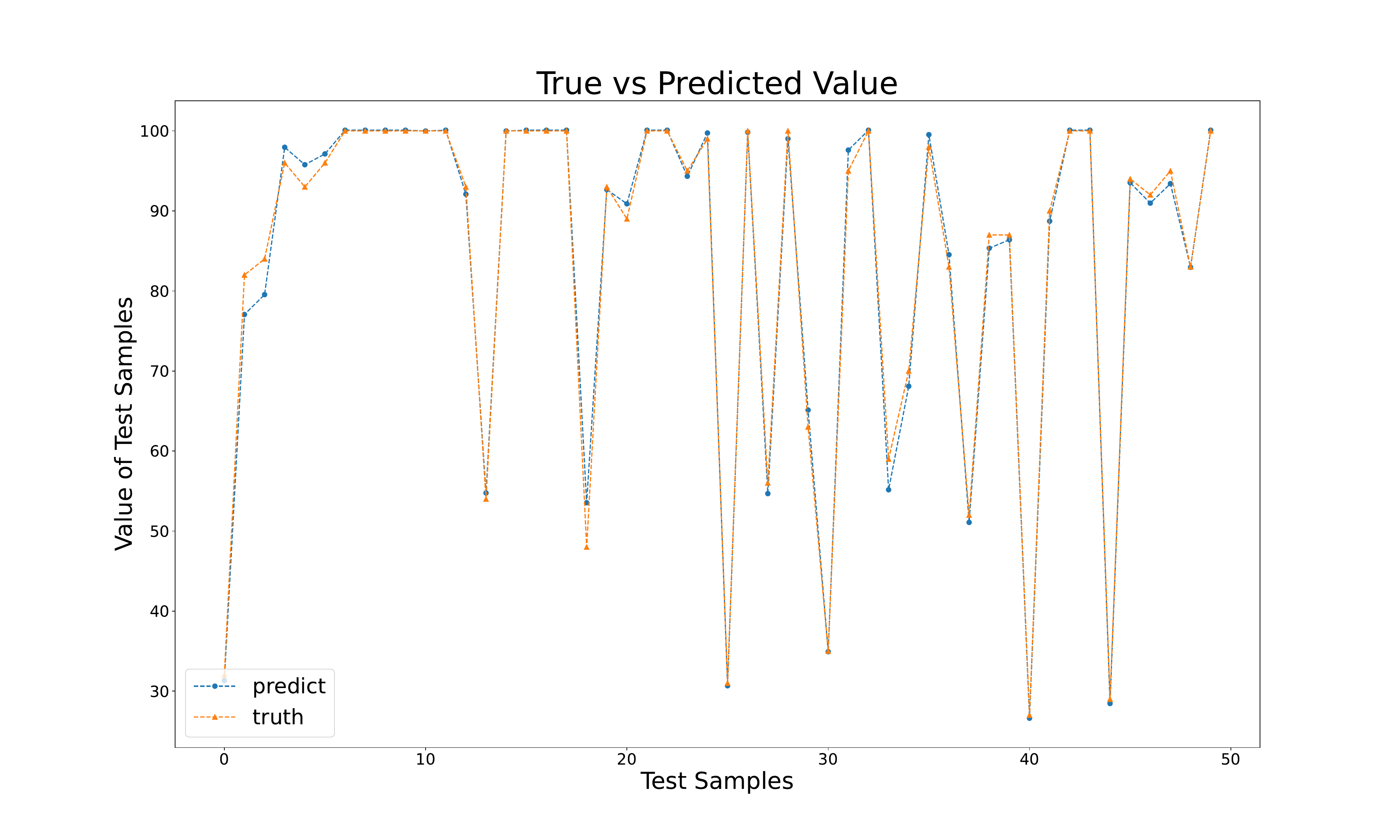}
    \caption{Learning coverage function with coverage probability 0.3, using the EDSF architecture, showing Training loss, Truth vs. Predicted values for train and test samples}
    \label{fig:edsf-mid}
\end{figure}

\begin{figure}[ht]
    \centering
    \includegraphics[width=.6\textwidth]{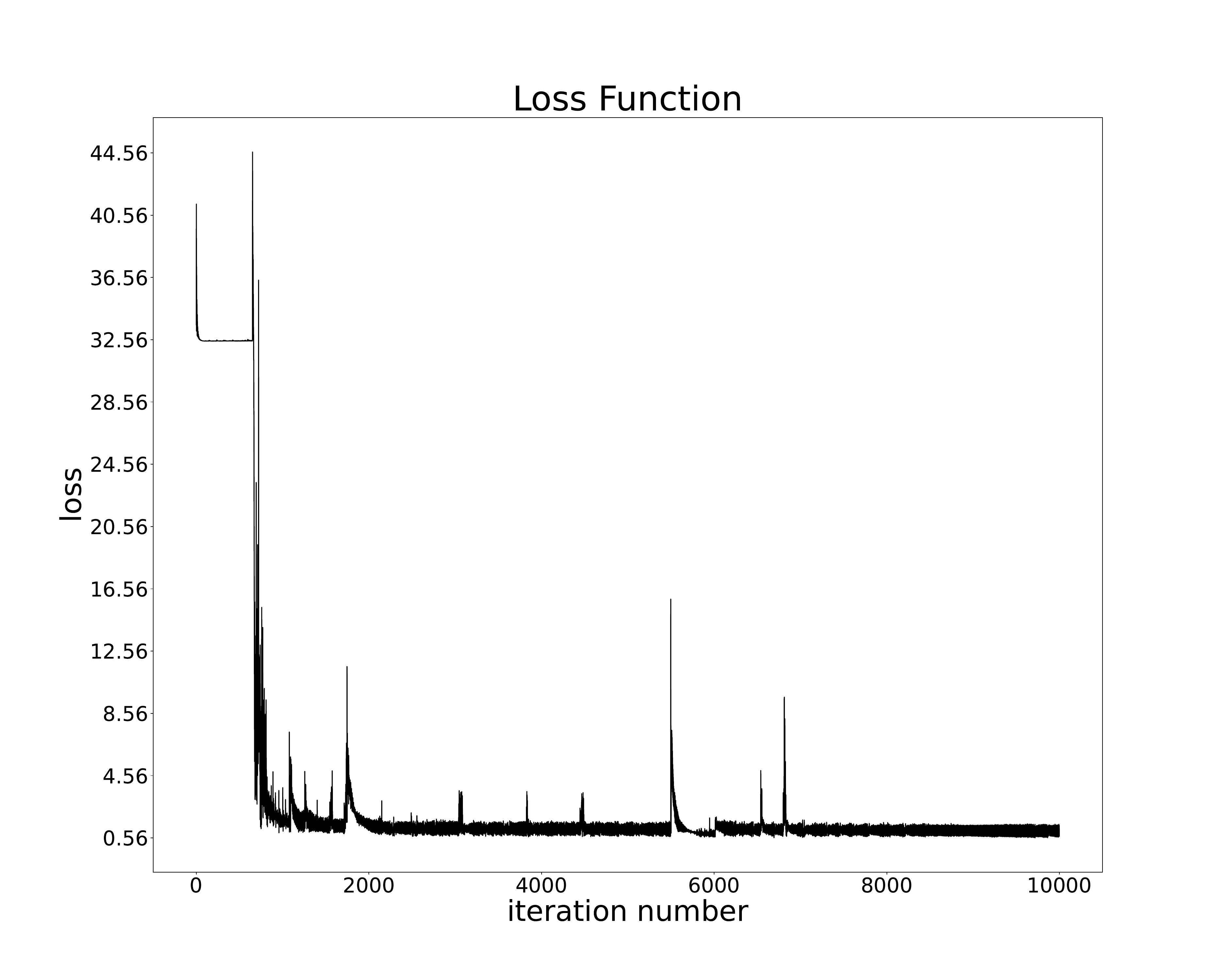}
    \includegraphics[width=.6\textwidth]{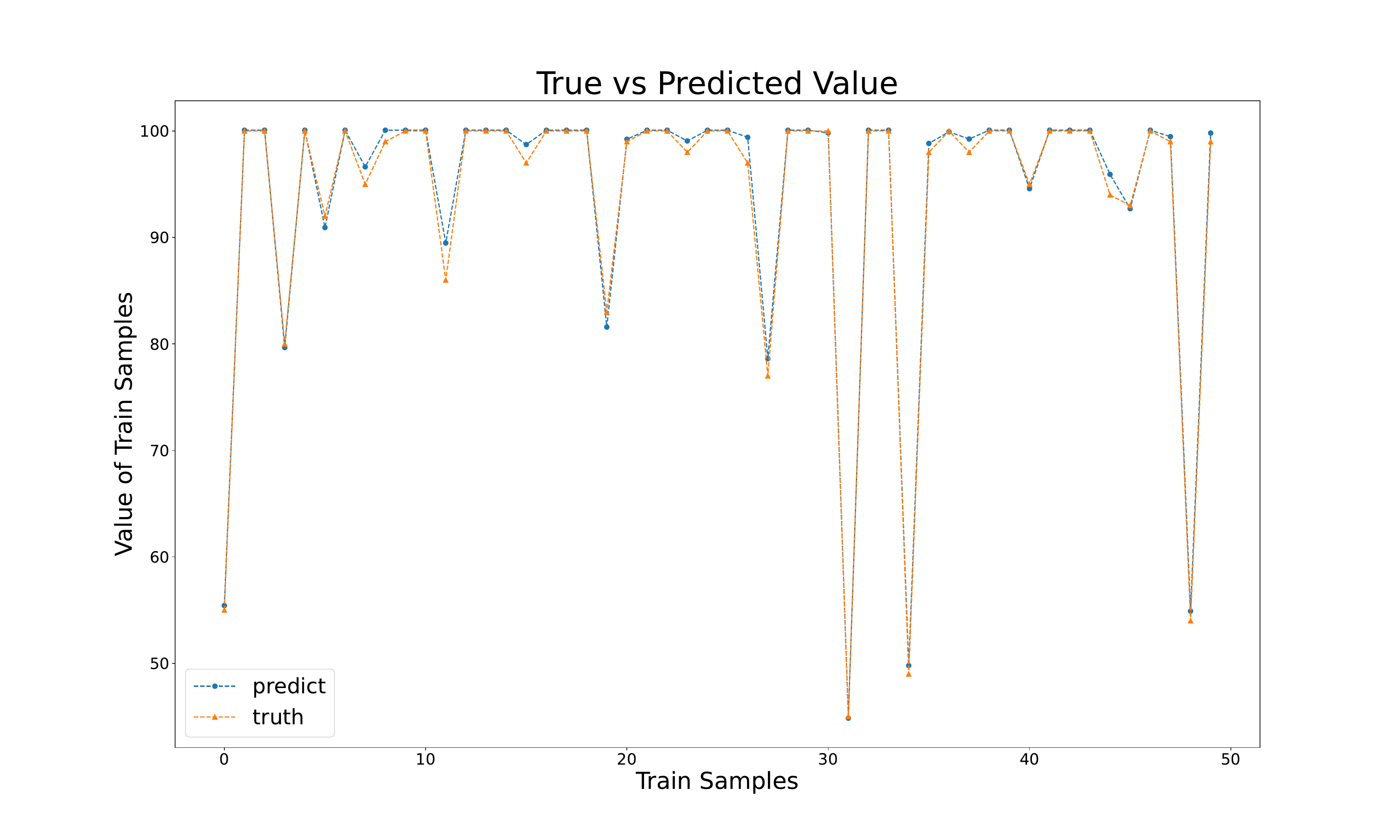}
    \includegraphics[width=.6\textwidth]{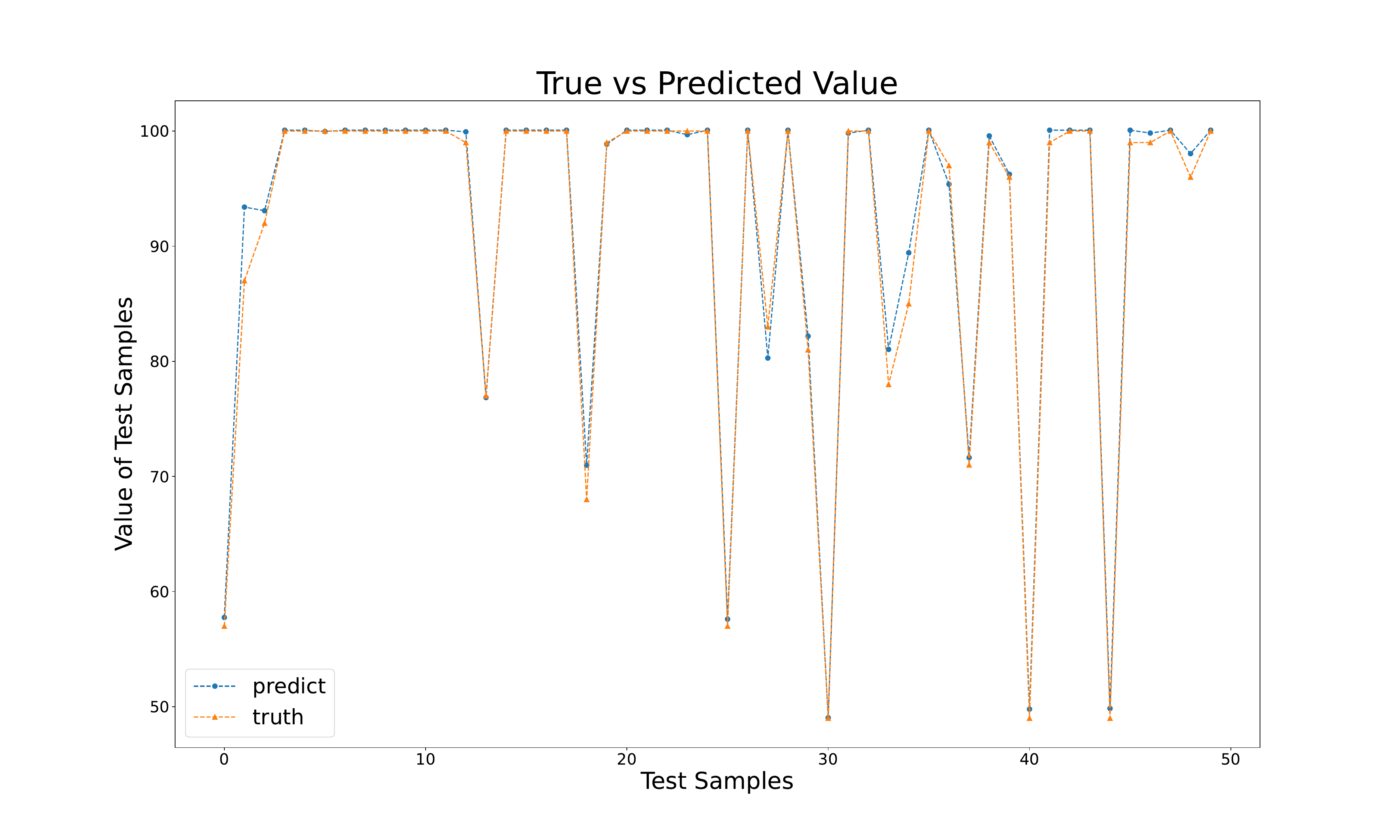}
    \caption{Learning coverage function with coverage probability 0.5, using the EDSF architecture, showing Training loss, Truth vs. Predicted values for train and test samples}
    \label{fig:edsf-high}
\end{figure}

\begin{figure}[ht]
    \centering
    \includegraphics[width=.6\textwidth]{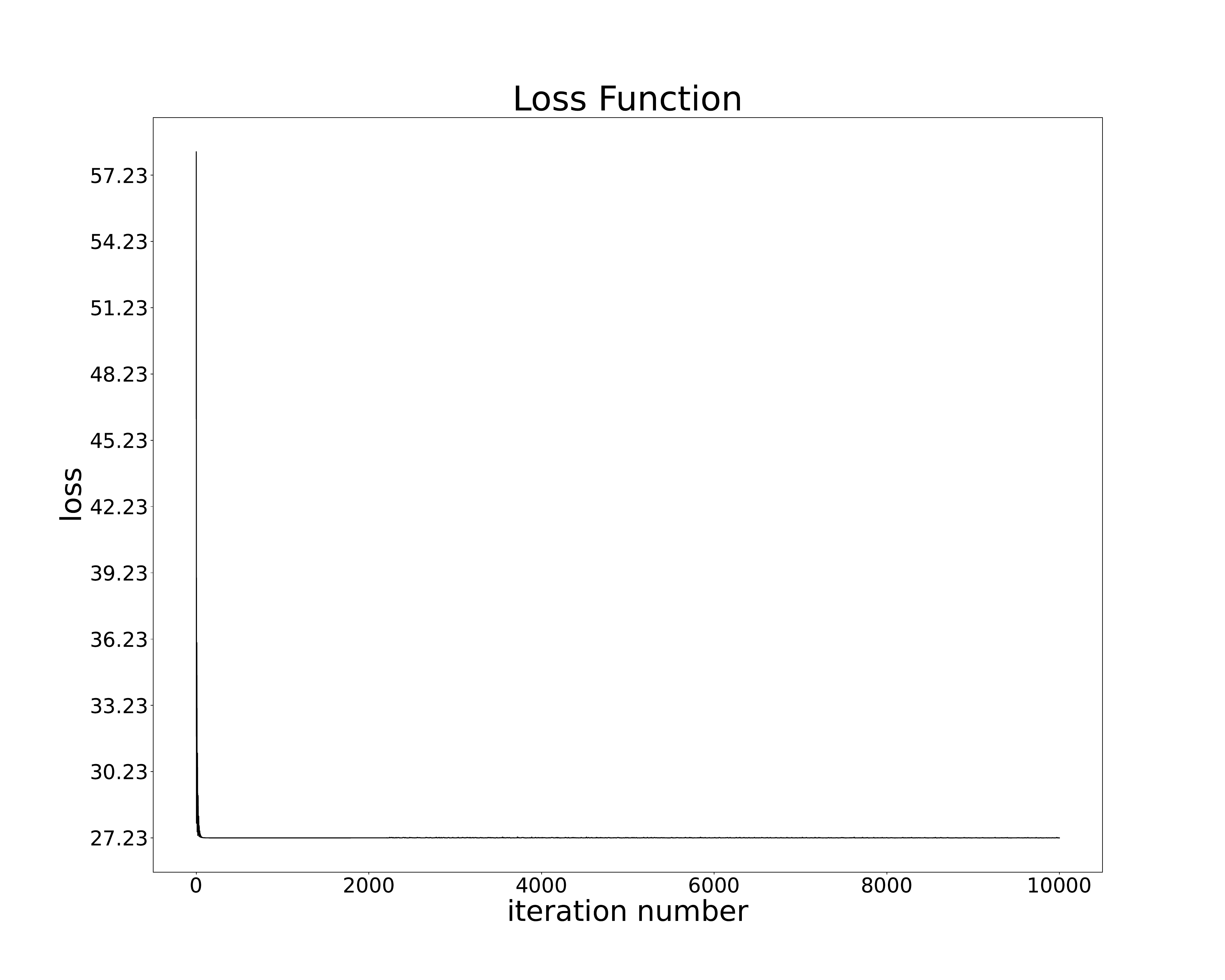}
    \includegraphics[width=.6\textwidth]{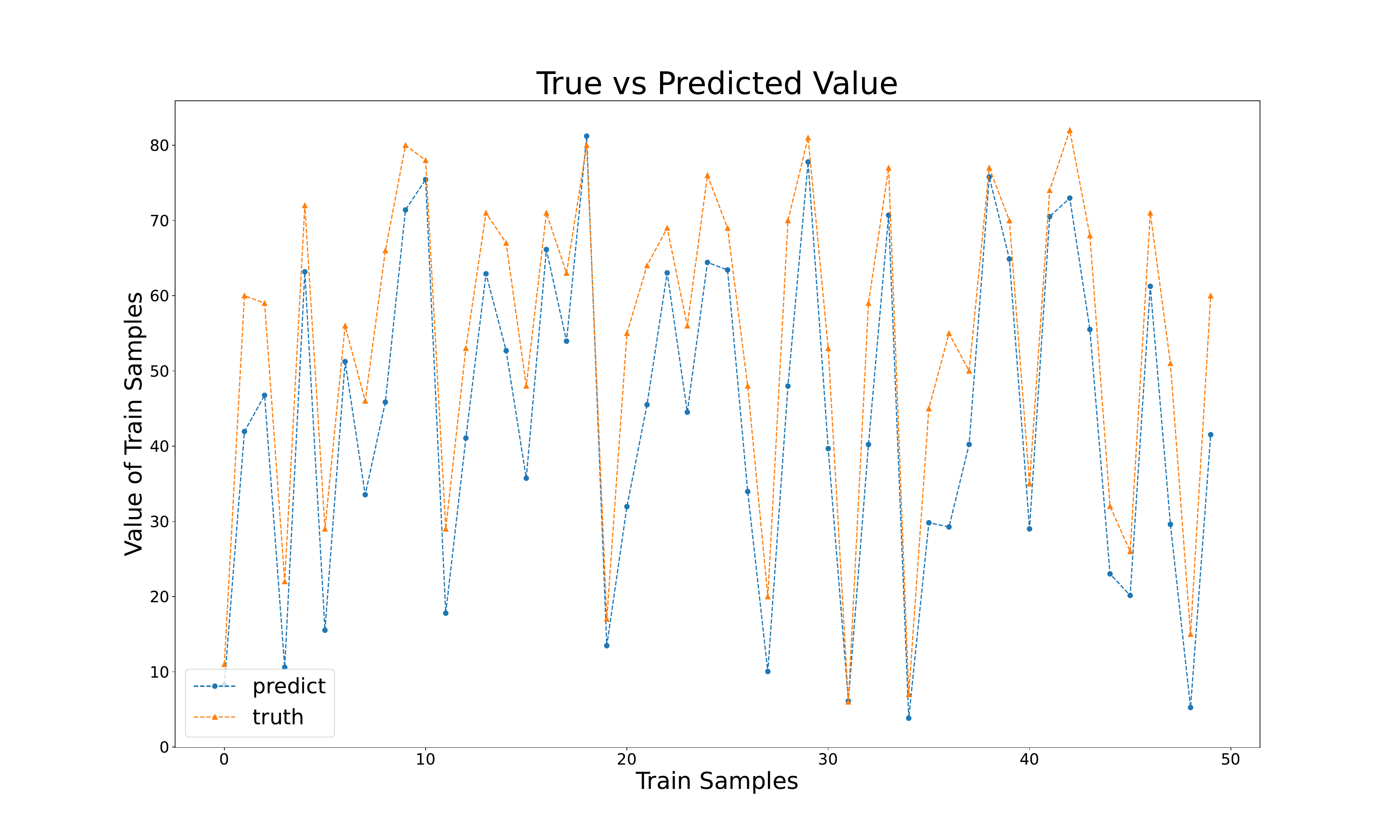}
    \includegraphics[width=.6\textwidth]{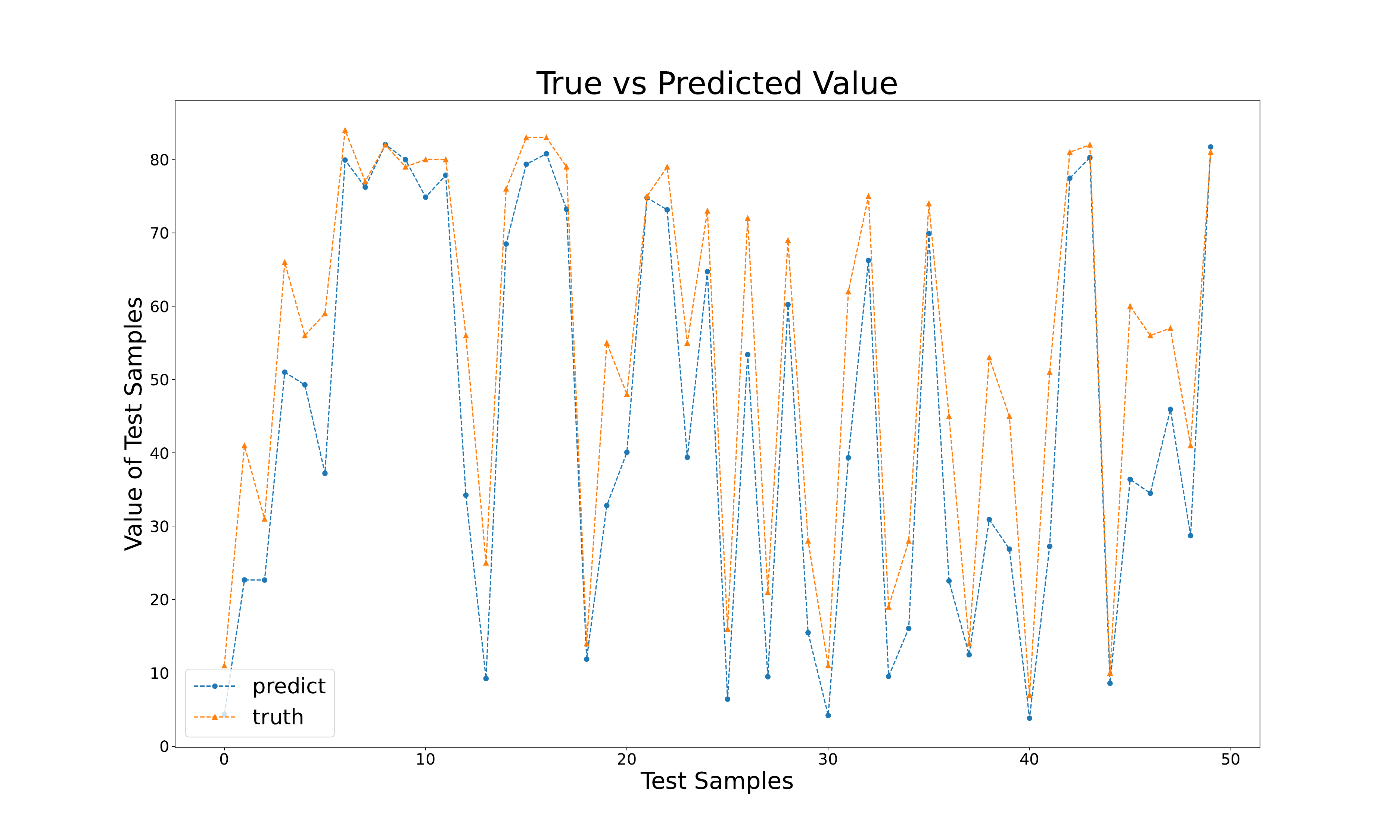}
    \caption{Learning coverage function with coverage probability 0.1, using the DSF architecture, showing Training loss, Truth vs. Predicted values for train and test samples}
    \label{fig:dsf-low}
\end{figure}

\begin{figure}[ht]
    \centering
    \includegraphics[width=.6\textwidth]{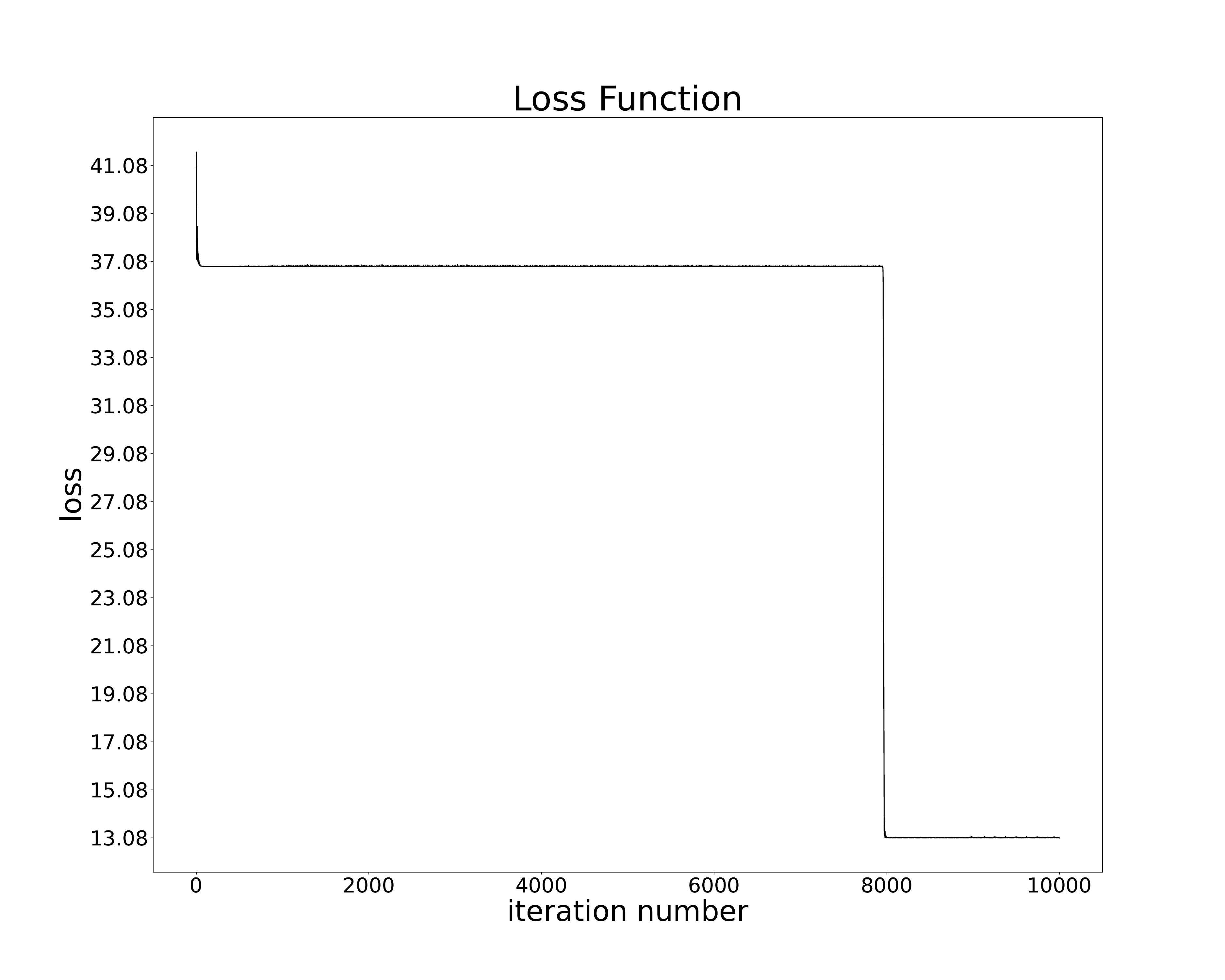}
    \includegraphics[width=.6\textwidth]{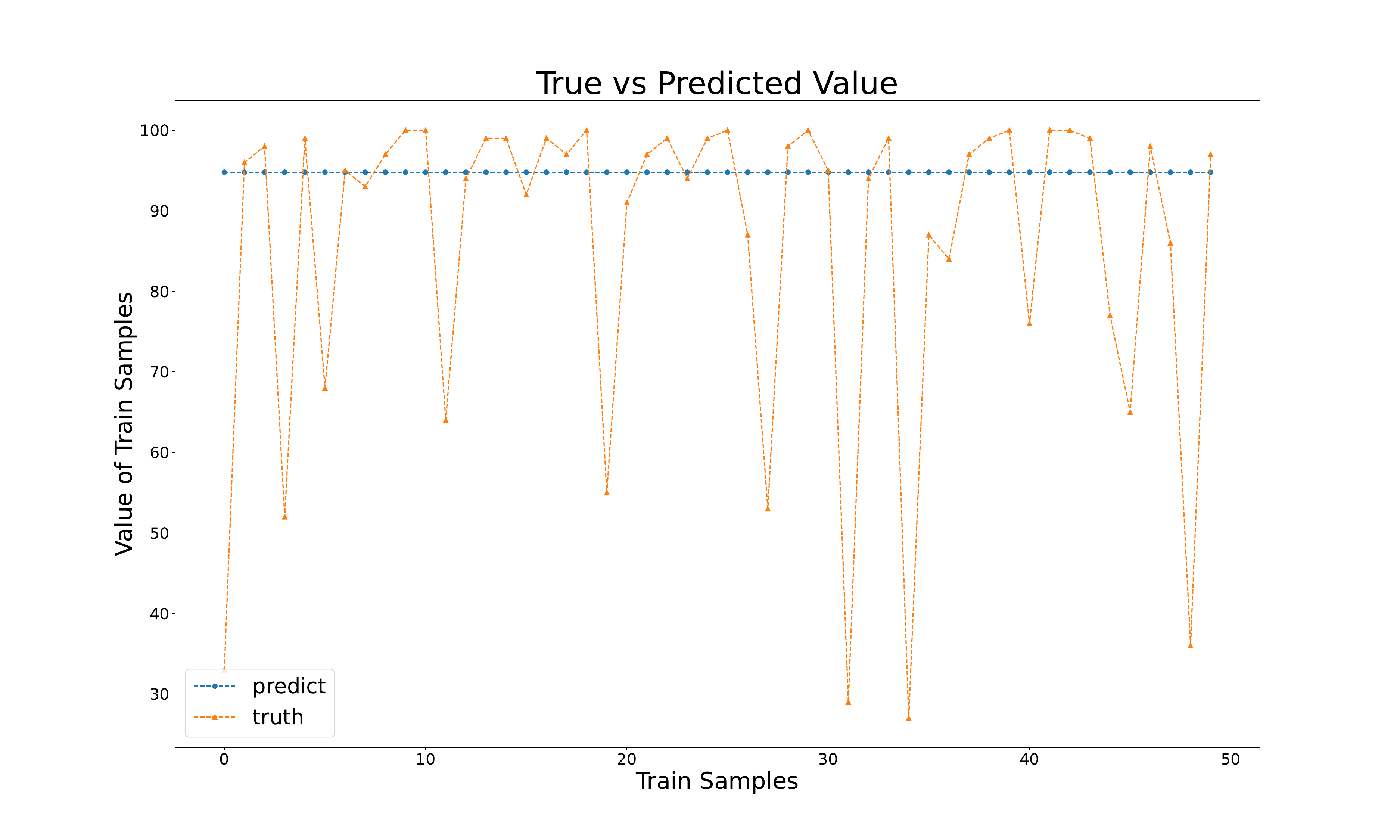}
    \includegraphics[width=.6\textwidth]{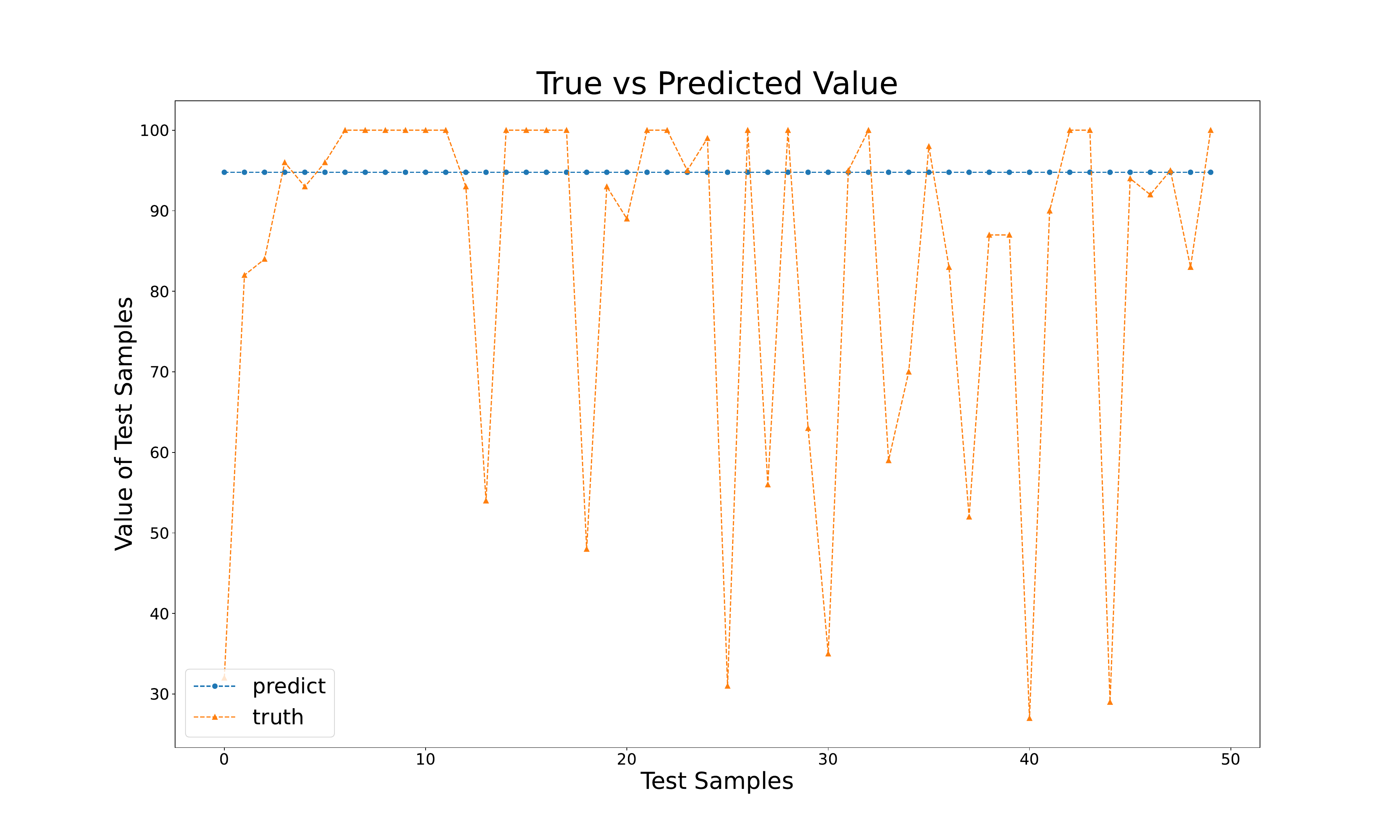}
    \caption{Learning coverage function with coverage probability 0.3, using the DSF architecture, showing Training loss, Truth vs. Predicted values for train and test samples}
    \label{fig:dsf-mid}
\end{figure}

\begin{figure}[ht]
    \centering
    \includegraphics[width=.6\textwidth]{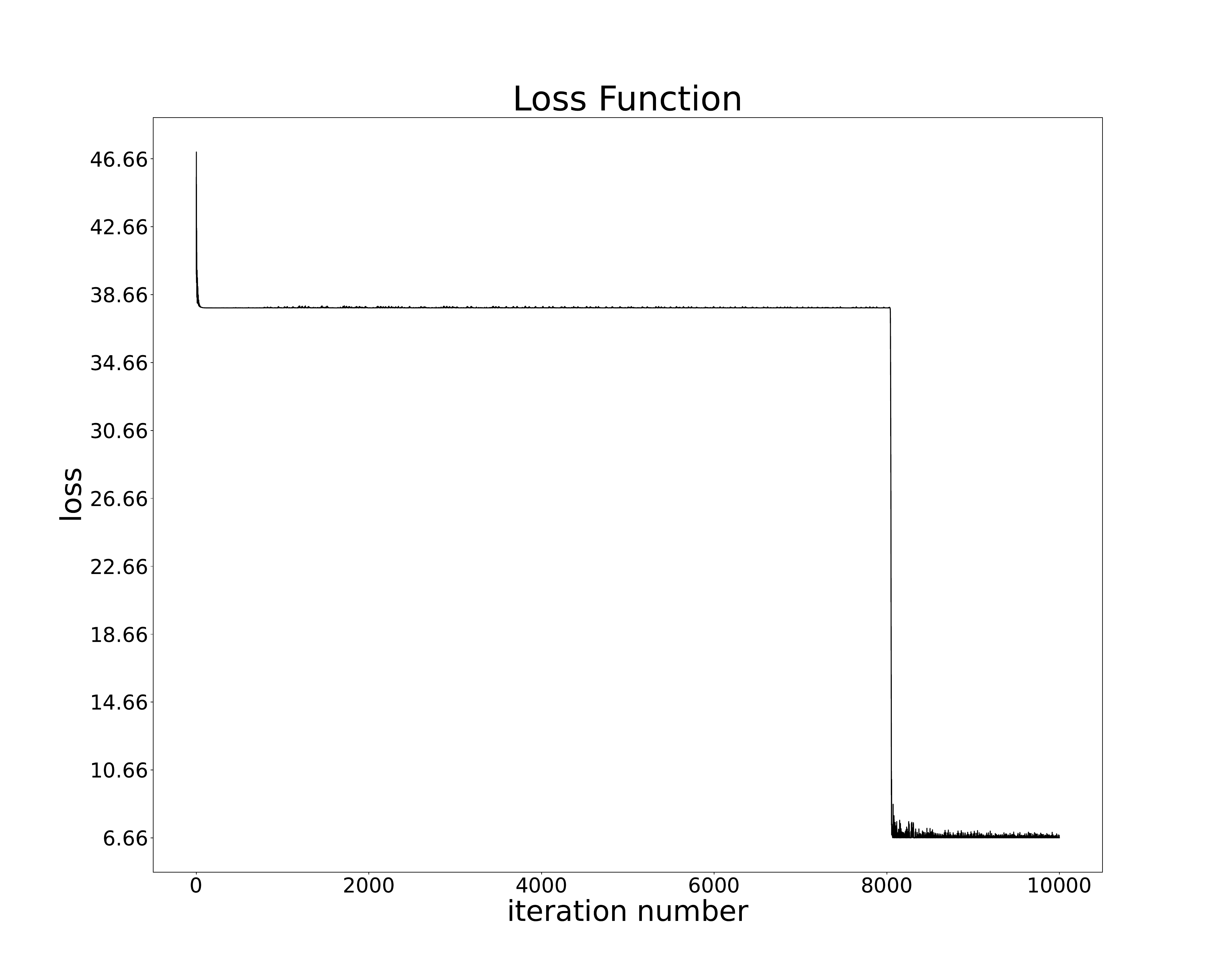}
    \includegraphics[width=.6\textwidth]{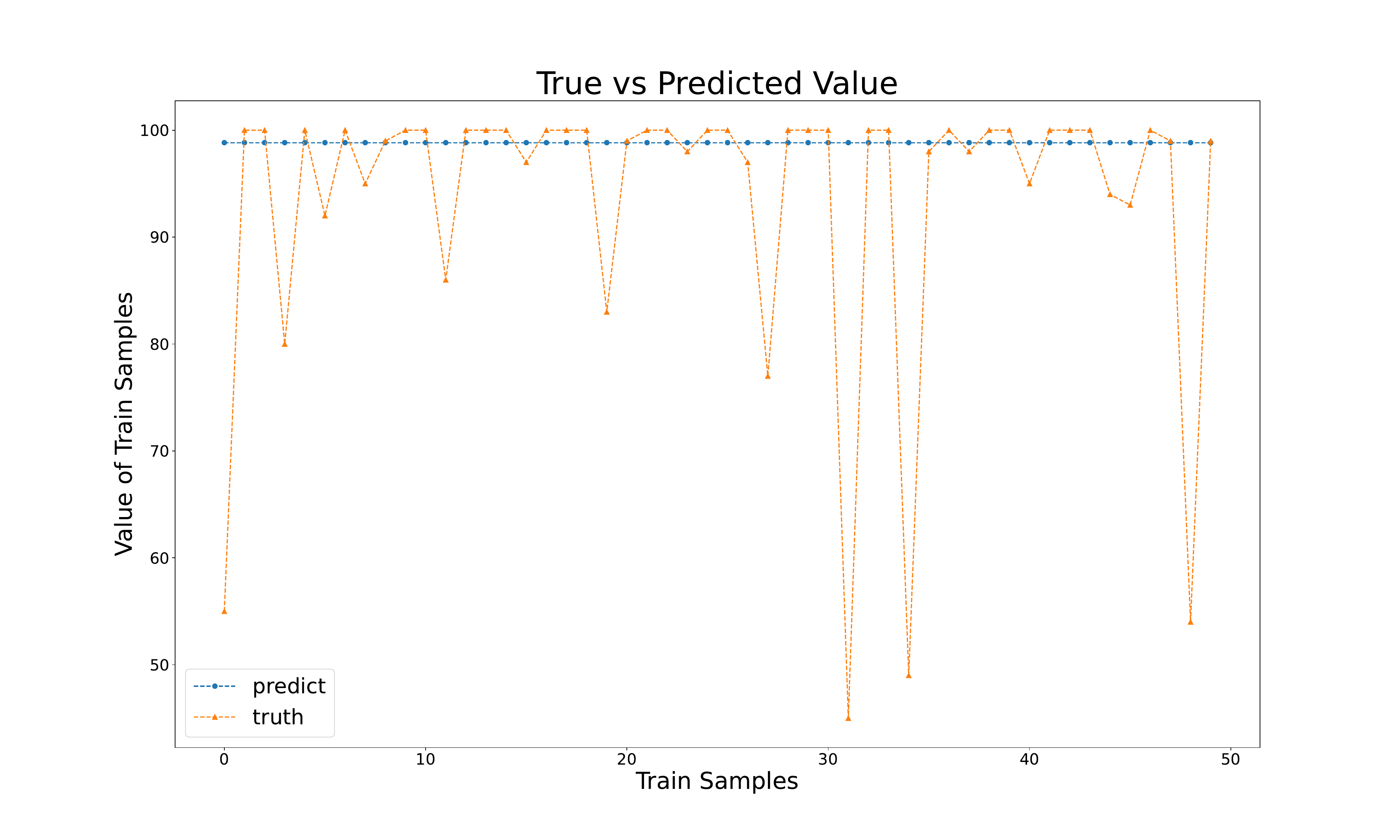}
    \includegraphics[width=.6\textwidth]{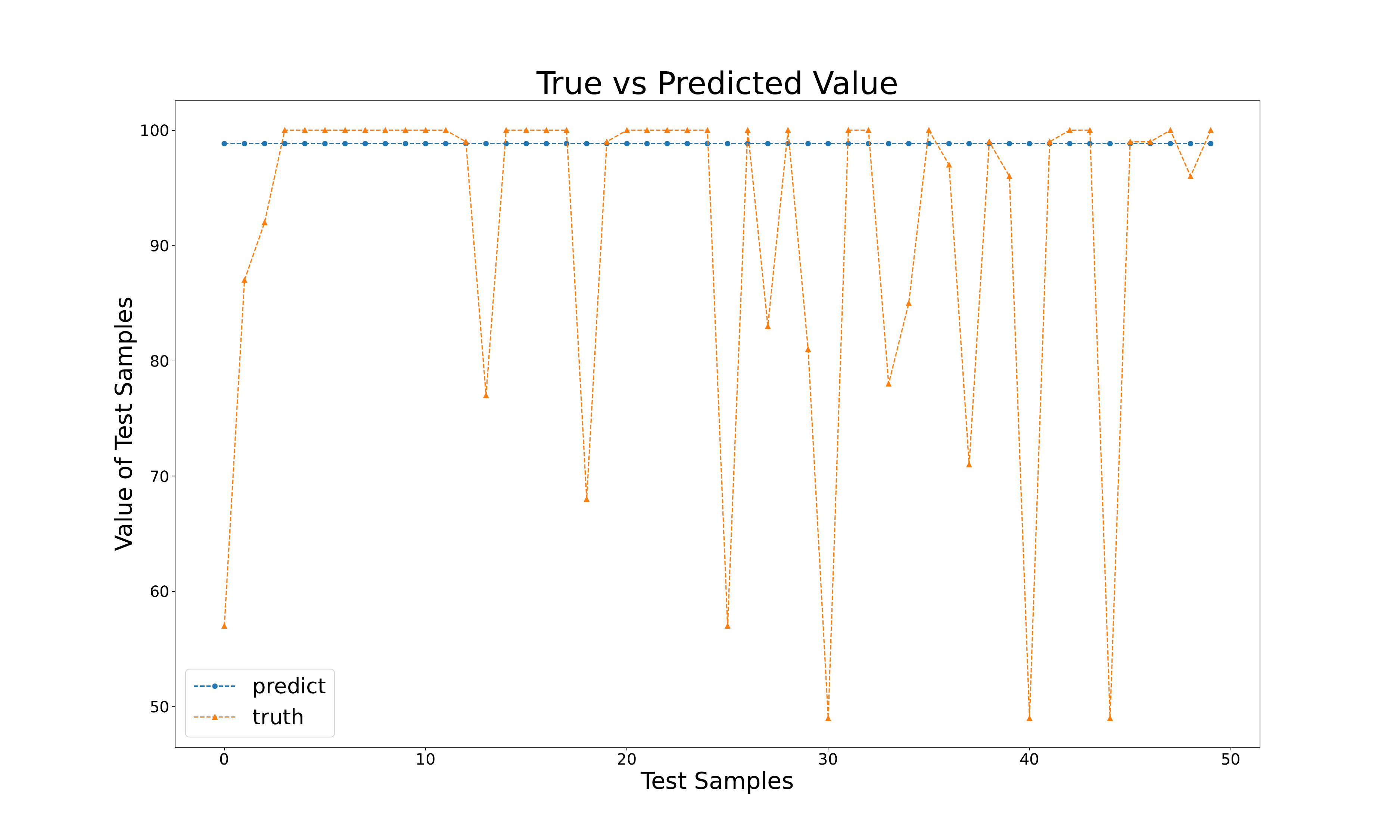}
    \caption{Learning coverage function with coverage probability 0.5, using the DSF architecture, showing Training loss, Truth vs. Predicted values for train and test samples}
    \label{fig:dsf-high}
\end{figure}

We have run each architecture for three distinct coverage functions described earlier in Section~\ref{sec:experimental_resutls}, and we have shown EDSFs has much lower empirical generalization error than DSFs. We illustrated this by providing a few samples' results, shown in Figures \ref{fig:edsf-low}-\ref{fig:dsf-high}. These figures represent the output of experiments conducted on generated data from each coverage function for EDSFs and DSFs separately. In each run we trained the above-mentioned neural networks on the data for 10000 epochs. As it can be seen, in EDSF experiments the predicted values completely follows the true values of the coverage function.

\section{Details of Experiments on Learning Cut Functions}\label{apndx:cut_function}
\begin{figure}[ht]
    \centering
    \includegraphics[width=.6\textwidth]{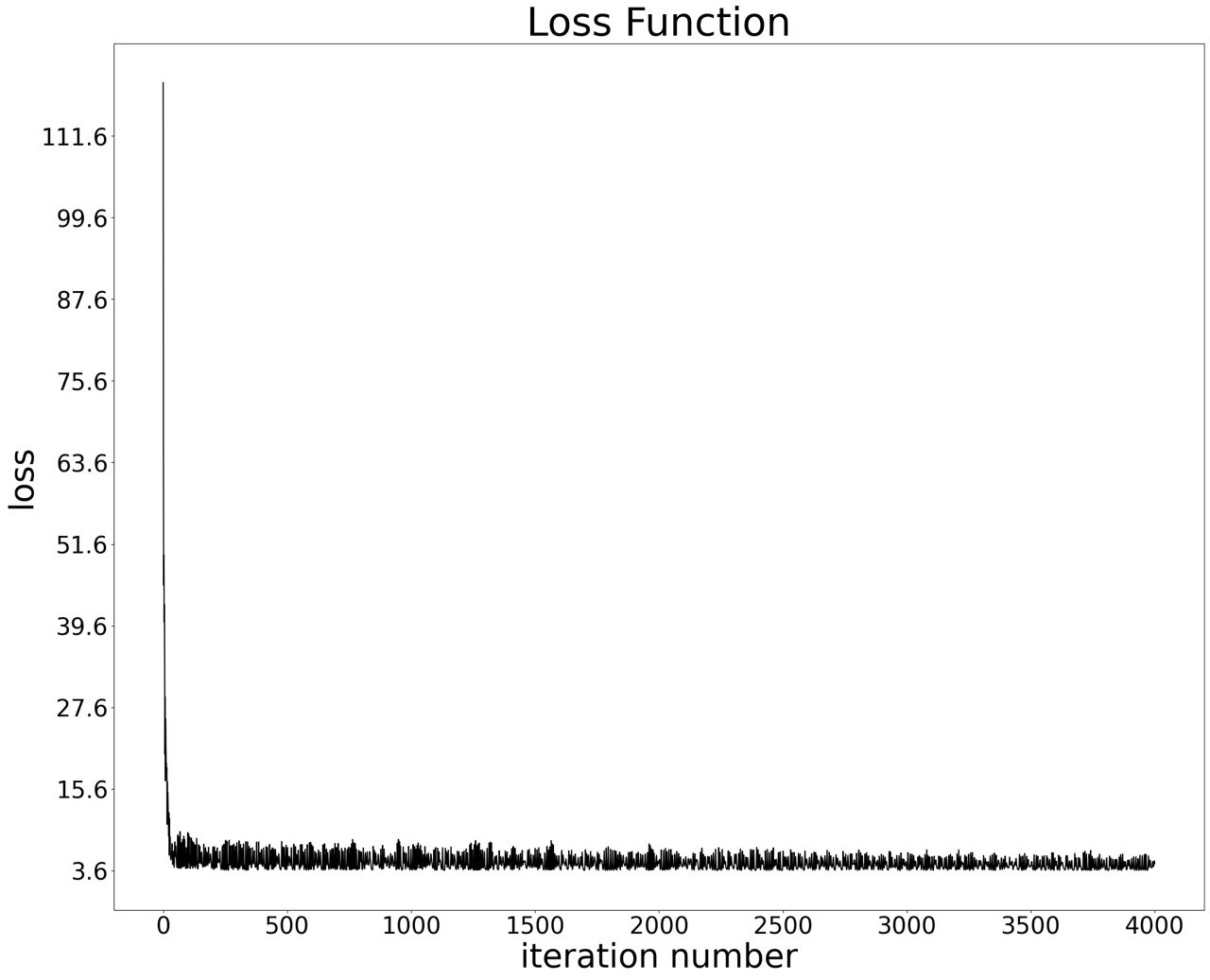}
    \includegraphics[width=.6\textwidth]{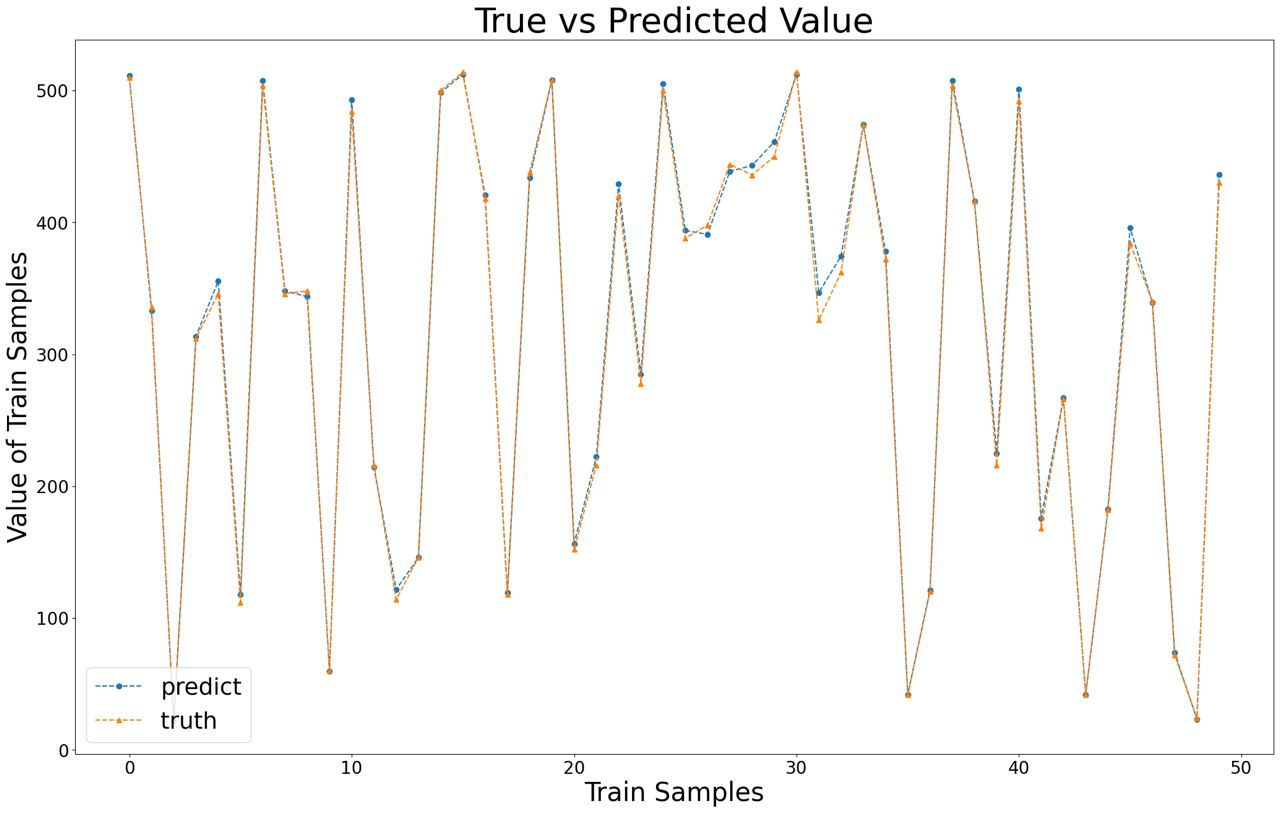}
    \includegraphics[width=.6\textwidth]{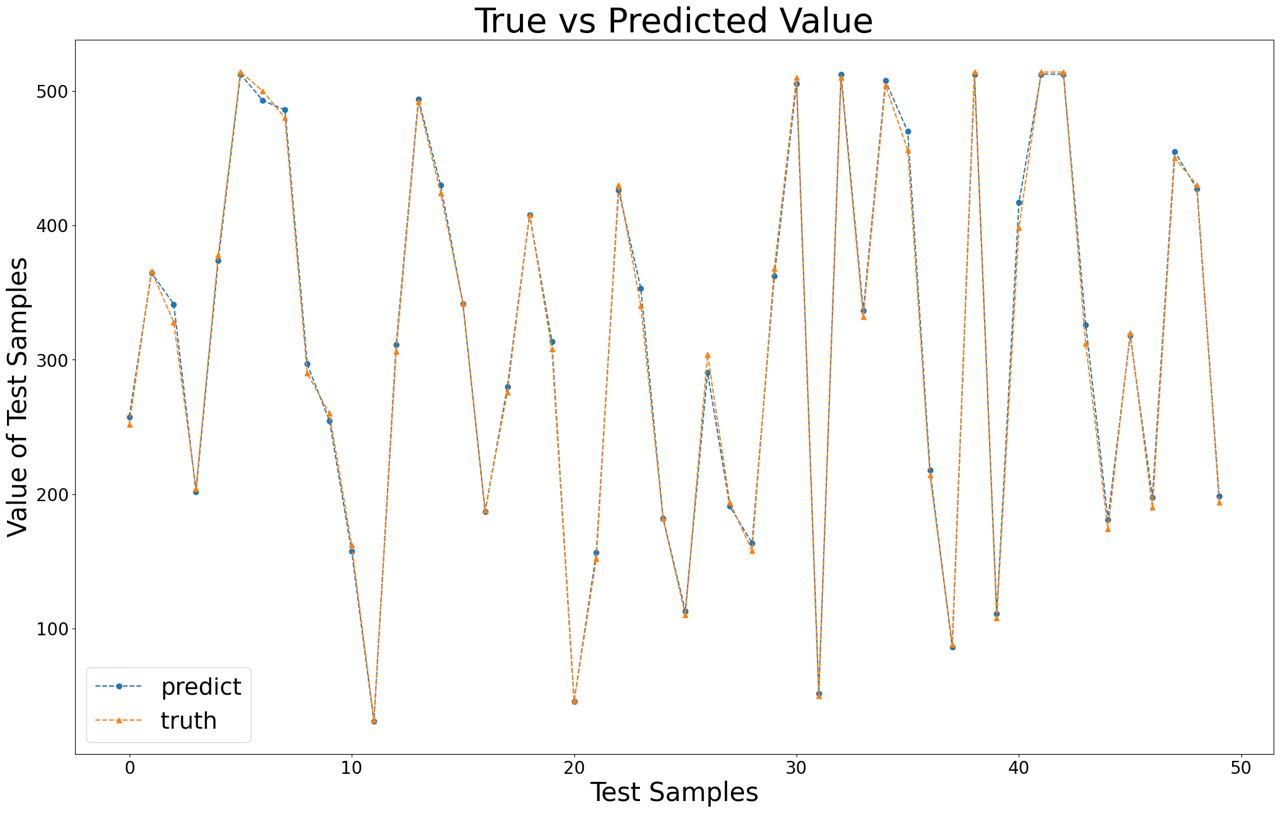}
    \caption{Learning a cut function generated by the Erdos-Renyi model with probability 0.2 and having 50 vertices, using the EDSF architecture, showing Training loss, Truth vs. Predicted values for train and test samples.}
    \label{fig:edsf-cut}
\end{figure}

In the learning cut function experiment, we have used the same architecture as the one used for learning coverage functions in Appendix~\ref{apndx:coverage_function}. We have presented the corresponding loss and generalization error for learning cut functions in the Figure \ref{fig:edsf-cut}.

\end{document}